# Most Relevant Explanation in Bayesian Networks

**Changhe Yuan**                                                    CYUAN@CSE.MSSTATE.EDU
*Department of Computer Science and Engineering*
*Mississippi State University*
*Mississippi State, MS 39762*

**Heejin Lim**                                                         HLIM@AI.KAIST.AC.KR
*Department of Computer Science*
*Korea Advanced Institute of Science and Technology*
*Daejeon, South Korea 305-701*

**Tsai-Ching Lu**                                                            TLU@HRL.COM
*HRL Laboratories LLC*
*Malibu, CA 90265*

## Abstract

A major inference task in Bayesian networks is explaining why some variables are observed in their particular states using a set of target variables. Existing methods for solving this problem often generate explanations that are either too simple (underspecified) or too complex (overspecified). In this paper, we introduce a method called *Most Relevant Explanation* (MRE) which finds a partial instantiation of the target variables that maximizes the *generalized Bayes factor* (GBF) as the best explanation for the given evidence. Our study shows that GBF has several theoretical properties that enable MRE to automatically identify the most relevant target variables in forming its explanation. In particular, *conditional Bayes factor* (CBF), defined as the GBF of a new explanation conditioned on an existing explanation, provides a soft measure on the degree of relevance of the variables in the new explanation in explaining the evidence given the existing explanation. As a result, MRE is able to automatically prune less relevant variables from its explanation. We also show that CBF is able to capture well the explaining-away phenomenon that is often represented in Bayesian networks. Moreover, we define two dominance relations between the candidate solutions and use the relations to generalize MRE to find a set of top explanations that is both diverse and representative. Case studies on several benchmark diagnostic Bayesian networks show that MRE is often able to find explanatory hypotheses that are not only precise but also concise.

## 1. Introduction

One essential quality of human experts is their ability to explain their reasoning to other people. In comparison, computer expert systems still lack the capability in that regard. Early medical decision-support systems such as MYCIN (Buchanan & Shortliffe, 1984) were shown empirically to have comparable or even better diagnostic accuracies than domain experts. However, physicians were still reluctant to use these systems in their daily clinical settings. One major reason is that these expert systems lack the capability to clearly explain their advice; physicians are uncomfortable in following a piece of advice that they do not fully understand (Teach & Shortliffe, 1981). The capability of explanation is thus critical for the success of a decision-support system.

Bayesian networks (Pearl, 1988) offer compact and intuitive graphical representations of the uncertain relations among the random variables in a domain and have become the basis of many probabilistic expert systems (Heckerman, Mamdani, & Wellman, 1995b). Bayesian networks provide principled approaches to finding explanations for given evidence, e.g., belief updating, Maximum a Posteriori assignment (MAP), and Most Probable Explanation (MPE) (Pearl, 1988). However, these methods may generate explanations that are either too simple (underspecified) or too complex (overspecified). Take a medical diagnostic system as an example. Such a system may contain dozens or even hundreds of potentially dependent diseases as target variables. *Target variables* are defined as the variables with diagnostic or explanatory interest. Belief updating finds only singleton explanations by ignoring the compound effect of multiple diseases. MAP and MPE consider such effect by finding a full configuration of the target variables, but their explanations often contain too many variables. Although a patient may have more than one disease, she almost never has more than a few diseases at one time as long as she does not delay treatments for too long. It is desirable to find explanations that only contain the most relevant diseases. Other diseases should be excluded from further medical tests or treatments.

In this paper, we introduce a method called *Most Relevant Explanation* (MRE) which finds a partial instantiation of the target variables that maximizes the *generalized Bayes factor* (GBF) as the best explanation for the given evidence. Our study shows that GBF has several theoretical properties that enable MRE to automatically identify the most relevant target variables in forming its explanation. In particular, *conditional Bayes factor* (CBF), defined as the GBF of a new explanation conditioned on an existing explanation, provides a soft measure on the degree of relevance of the variables in the new explanation in explaining the evidence given the existing explanation. As a result, MRE is able to automatically prune less relevant variables from its explanation. We also show that CBF is able to capture well the explaining-away phenomenon that is often represented in Bayesian networks. Moreover, we define two dominance relations between the candidate solutions and use the relations to generalize MRE to find a set of top explanations that is both diverse and representative. Our case studies show that MRE performed well on the explanation tasks in a set of benchmark Bayesian networks.

The remainder of the paper is structured as follows. Section 2 provides a brief overview of the literature on explanation, including scientific explanation, explanation in artificial intelligence, and the relation between causation and explanation. Section 3 provides an introduction to explanation in Bayesian networks, especially methods for explaining evidence. Section 4 introduces the formulation of *Most Relevant Explanation* and discusses its theoretical properties. This section also discusses how to generalize MRE to find a set of top explanations. Section 5 presents the case studies of MRE on a set of benchmark Bayesian networks. Finally, Section 6 concludes the paper.

## 2. Explanation

Explanation is a topic that is full of debate; in fact, there is no commonly accepted definition of explanation yet. This section provides a very brief overview of the major developments of explanation in both philosophy of science and artificial intelligence. A brief discussion on the relation between causation and explanation is also included.

## 2.1 Scientific Explanation

Explanation has been the focal subject of philosophy of science for a long time. The goal of explanation is not simply to describe the world as it is, but to develop a fundamental and scientific understanding of the world (Woodward, 2003) and answer questions such as "why did this happen?" The field is hence named *scientific explanation*.

One of the earliest model of scientific explanation is the Deductive-Nomological (D-N) model (Hempel & Oppenheim, 1948). According to the D-N model, a scientific explanation consists of an explanandum, the phenomenon to be explained, and an explanation (or explanans), the facts used to explain the explanandum. The explanation can successfully explain the explanandum if the explanation is true, and the explanation logically entails the explanandum.

However, not every phenomenon can be expressed in terms of deterministic logic rules. Many phenomena are inherently uncertain. Hempel (1965) modified his D-N model and introduced the *inductive-statistical* (I-S) model. The I-S model is similar to the D-N model except that it assumes there is a probability for each logic rule for capturing the uncertainty in linking the initial conditions to the phenomenon to be explained.

Both the D-N and I-S models have the limitation of allowing the inclusion of irrelevant facts in an explanation, because such facts do not affect the correctness of the rules (Suermondt, 1992). To address the shortcoming, Salmon (1970) introduced the *statistical-relevance* (S-R) model. The S-R model requires that an explanation should only consist of facts that are statistically relevant to the explanandum. A fact is statistically relevant in explaining the explanandum if the posterior probability of the fact after observing the explanandum is different from its prior probability. The intuition behind the S-R model is that statistically irrelevant facts, even though they may have high probabilities, do not constitute a good explanation.

Salmon (1984) introduced the *Causal-Mechanical* (C-M) model of explanation to take into account causation. The basic idea behind the C-M model is that the process of explanation involves fitting an explanandum into a causal structure of a domain and tracing the causes that may lead to the explanandum.

There are many other approaches as well. Ketcher (1989) believes that a scientific explanation should provide a unified account of the natural phenomena in the world. Van Fraassen (1980) believes that an explanation should favor the explanandum, i.e., the explanation either increases the probability of the explanandum or decreases the probability of the nearest competitor of the explanandum. Several mathematical theories of explanatory power have also been proposed by Jeffreys (1935), Good (1977), and Gärdenfors (1988).

## 2.2 Explanation in Artificial Intelligence

In comparison to scientific explanation, researchers in the area of artificial intelligence have taken a much broader view of explanation. Early development of decision-support systems made it clear that decision-support systems should not be intended to replace human experts, but rather provide advice or second opinion to the experts so that they can have better performance. So it is necessary that the decision-support systems have the capability to explain how their conclusions are made and why the conclusions are appropriate so that the domain experts can understand and possibly follow the advice (Dannenberg, Shapiro,

& Fries, 1979). In a decision-support system, any presentation that can help a user's understanding of the conclusions of the system is regarded as an explanation (Suermondt, 1992). For example, an explanation can be a trace of the reasoning process of a system in reaching its conclusions (Suermondt, 1992), a canned textual explanation of an abstract reasoning process (Bleich, 1972), a verbal translation of the inference rules (Buchanan & Shortliffe, 1984), or a visual display of certain elements of the system that helps a user to understand the results (Lacave, Luque, & Diez, 2007).

Nevertheless, the most important form of explanation in artificial intelligence is the forming of explanatory hypotheses for observed facts, often called *abductive inference* (Peirce, 1948). Many issues need consideration in abductive inference. One issue is to define what to explain. Not every observation needs explanation. According to Pierce (1948), a common trigger for abductive inference is that a *surprising* fact is observed, that is, the newly observed information is in conflict with what is already known or what is currently believed. In other words, what requires explanation is something that causes someone's *cognitive dissonance* between the explanandum and the rest of her belief (Gärdenfors, 1988). It is also argued that some observations can serve as part of an explanation (Chajewska & Halpern, 1997; Nielsen, Pellet, & Elisseeff, 2008). Suppose we observe that the grass is wet, and it rained. There is no need to explain why it rained. The fact that it rained is actually an excellent explanation for why the grass is wet (Nielsen et al., 2008). Therefore, there is a potential distinction between the *explanandum*, i.e., observations to be explained, and the other observations. Deciding the explanandum may be a nontrivial task.

Once an explanandum is decided, the task reduces to finding an explanatory hypothesis for the explanandum. The issue here is to define what a good explanation is. A good explanation should be able to provide some sort of *cognitive relief*, that is, the explanation should decrease the surprise value caused by the observation of the explanandum. Intuitively, the value of an explanation is the degree to which the explanation decreases the surprise value. Many different criteria have been used in existing explanation methods, including weight of evidence (Good, 1985), probability (Pearl, 1988), explanatory power (Gärdenfors, 1988), likelihood of evidence (de Campos, Gamez, & Moral, 2001), and causal information flow (Nielsen et al., 2008). One goal of this paper is to study and compare the properties of some of these measures.

Moreover, the quality of an explanation is also highly goal-dependent. The resulting explanations may vary in the level of specificity or scope depending on the objectives of the explanation task (Leake, 1995). For example, when explaining the symptoms of a patient, a doctor can either find an explanation that constitutes a diagnosis, i.e., explaining which diseases the patient may have, or provide an explanation on which risk factors may have caused the symptoms. Defining the goal of an explanation task is thus important.

The above is a general overview of explanation in artificial intelligence. In Section 3, we will provide a more detailed review of one particular topic in this area: explanation in Bayesian networks.

## 2.3 Causation and Explanation

It is agreed upon that causation plays an important role in explanation and helps to generate intuitive explanations (Chajewska & Halpern, 1997; Halpern & Pearl, 2005; Nielsen et al., 2008). However, there is considerable disagreement among researchers about whether all explanations are causal and what the distinction between causal and non-causal explanations is. Those who believe in not completely subsuming explanation into causation typically have a broader view of explanation. The following are some arguments for the belief that there are explanation tasks that do not require causal meanings.

First of all, explanation has a much broader meaning in AI. Some explanation tasks in AI do not require causal meanings. For example, verbal or visual explanations are often used in decision-support systems to illustrate concepts, knowledge, or reasoning processes; they do not seem to have causal meanings. If it is insisted that all explanations have to be causal, all these explanation tasks may have to be disallowed.

Furthermore, there are many domains in which clear understandings of causal relations have not yet been established. There is only statistical relevance information. Such relevance information is insufficient to fully capture causal relations. Again, explanation may have to be disallowed altogether in these domains if we insist that all explanations have to be causal.

Moreover, the situation is further complicated by the fact that there is still considerable disagreement on the definition of causation. Causal claims themselves also vary drastically in the extent to which they are explanatorily deep enough (Woodward, 2003). For some, it is a sufficient causal explanation to say that the rock broke the window. For others, that is merely a descriptive statement; it is necessary to resort to deeper Newtonian mechanics to explain why the rock broke the window.

Finally, not all legitimate why-questions are causal or require causal explanations. For example, there is a difference between the explanation of belief and the explanation of fact (Chajewska & Halpern, 1997). When someone tells you that it rained last night and you ask her why, she may give you as an explanation that the grass is wet. Another example is that a variety of physical explanations are geometrical rather than causal because they explain phenomena by using the structure of spacetime rather than using forces or energy transfer (Nerlich, 1979).

It is not our goal in this paper to take on the tall task of settling the debate on the relation between causation and explanation. The methods that we propose in this paper are aimed to be general enough so that they are applicable to both causal and non-causal settings.

## 3. Explanation in Bayesian Networks

A Bayesian network is a directed acyclic graph (DAG) in which the nodes denote random/chance variables, and the arcs or lack of them denote the qualitative relations among the variables. The dependence relations between the variables are further quantified with conditional probability distributions, one for each variable conditioned on its parents. Figure 1(b) shows an example of a Bayesian network. A Bayesian network essentially encodes a joint probability distribution over the random variables of a domain and can serve as a probabilistic expert system to answer various queries about the domain.

Unlike many machine learning methods that are mostly predictive methods, a Bayesian network can be used for both prediction and explanation with a deep representation of a domain. Explanation tasks in Bayesian networks can be classified into three categories;

explanation of reasoning, explanation of model, and explanation of evidence (Lacave & Diez, 2002). The goal of *explanation of reasoning* in Bayesian networks is to explain the reasoning process used to produce the results so that the credibility of the results can be established. Because reasoning in Bayesian networks follows a normative approach, explanation of reasoning is more difficult than explanation in methods that try to imitate human reasoning (Druzdzel, 1996; Lacave & Diez, 2002). The goal of *explanation of model* is to present the knowledge encoded in a Bayesian network in easily understandable forms such as visual aids so that experts or users can examine or even update the knowledge. For example, graphical modeling tools such as GeNIe (Druzdzel, 1999), Elvira (Lacave et al., 2007), and SamIam (AR Group, UCLA, 2010) have functionalities for visualizing the strength of the probabilistic relations between the variables in a domain. The goal of *explanation of evidence* is to explain why some observed variables are in their particular states using the other variables in the domain. We focus on the explanation of evidence in this research. Next we review the major methods for explaining evidence in Bayesian networks and discuss their limitations.

## 3.1 Explanation of Evidence

Numerous methods have been developed to explain evidence in Bayesian networks. Some of these methods make simplifying assumptions and focus on singleton explanations. For example, it is often assumed that the fault variables are mutually exclusive and collectively exhaustive, and there is conditional independence of evidence given any hypothesis (Heckerman, Breese, & Rommelse, 1995a; Jensen & Liang, 1994; Kalagnanam & Henrion, 1988). However, singleton explanations may be *underspecified* and are unable to fully explain the given evidence if the evidence is the compound effect of multiple causes.

For a domain with multiple dependent target variables, multivariate explanations are often more appropriate for explaining the given evidence. *Maximum a Posteriori assignment* (MAP) finds a complete instantiation of a set of target variables that maximizes the joint posterior probability given partial evidence on the other variables. *Most Probable Explanation* (MPE) (Pearl, 1988) is similar to MAP except that MPE defines the target variables to be all the unobserved variables. The common drawback of these methods is that they often produce hypotheses that are *overspecified* and may contain irrelevant variables in explaining the given evidence.

Everyday explanations are necessarily partial explanations (Leake, 1995). It is difficult and also unnecessary to account for all the potential factors that may be related to the occurrence of an event; it is desirable to find the *most relevant* contributing factors. Various pruning techniques have been used to avoid overly complex explanations. These methods can be grouped into two categories: *pre-pruning* and *post-pruning*. *Pre-pruning* methods use the context-specific independence relations represented in Bayesian networks to prune irrelevant variables (Pearl, 1988; Shimony, 1993; van der Gaag & Wessels, 1993, 1995) before applying methods such as MAP to generate explanations. For example, Shimony (1993) defines an explanation as the most probable independence-based assignment that is complete and consistent with respect to the evidence nodes. Roughly speaking, an explanation is a truth assignment to the variables relevant to the evidence nodes. Only the ancestors of the evidence nodes can be relevant. An ancestor of a given node is irrelevant if it is independent of that node given the values of other ancestors. However, these independence relations are too strict and are unable to prune marginally or loosely relevant target variables.

In contrast, *post-pruning* methods first generate explanations using methods such as MAP or MPE and then prune variables that are not important. An example is the method proposed by de Campos et al. (2001). Their method first finds the K most probable explanations (K-MPE) for the given evidence, where K is a user-specified parameter, and then simplify the explanations by removing unimportant variables one at a time. A variable is regarded as unimportant if its removal does not reduce the likelihood of an explanation.

It is pointed out that Shimony's partial explanations are not necessarily concise (Chajewska & Halpern, 1997). His explanation must include an assignment to all the nodes in at least one path from a variable to the root, since for each relevant node, at least one of its parents must be relevant. The method by de Campos et al. can only prune conditionally independent variables as well. The limitations of both methods were addressed by using a thresholding method to allow the pruning of marginally relevant variables (Shimony, 1996; de Campos et al., 2001). However, the modified methods involve the manual setting of some tunable parameters, which can be rather arbitrary and are subject to human errors.

Several methods use the likelihood of evidence to measure the explanatory power of an explanation (Gärdenfors, 1988). Chajewska and Halpern (1997) extend the approach further to use the value pair of <likelihood, prior probability> to order the explanations, forcing users to make decisions if there is no clear order between two explanations. Since the likelihood measure allows comparing explanations that contain different numbers of variables, it can potentially find more concise explanations. But in practice these methods often fail to prune irrelevant variables, because adding such variables typically does not affect the likelihood.

Henrion and Druzdzel (1991) assume that a system has a set of pre-defined explanation scenarios organized as a tree; they use the scenario with the highest posterior probability as the explanation. This method also allows comparing explanations with different numbers of variables but requires the explanation scenarios to be specified in advance.

Flores et al. (2005) propose to automatically create an explanation tree by greedily branching on the most informative variable at each step while maintaining the probability of each branch of the tree above a certain threshold. It is pointed out that Flores et al.'s approach adds variables in the order of how informative they are about the remaining target variables, not how informative they are about the explanandum (Nielsen et al., 2008). The results in this paper provide evidence for that drawback as well. Moreover, the criterion that they use to choose the best explanation is the probability of the explanation given the evidence, which makes the ranking of the explanations extremely sensitive to a user-specified threshold for bounding the probabilities of the branches. Nielsen et al. developed another method that uses the *causal information flow* (Ay & Polani, 2008) to select variables to expand an explanation tree. However, it inherits the drawback that explanation tree-based methods are in essence greedy search methods; even though they can identify important individual variables, they may fail to recognize the compound effect of multiple variables. Furthermore, the method also has tunable parameters that may be subject to human errors. Finally, since each explanation in an explanation tree has to contain a full branch starting from the root, such explanations may still contain redundant variables.

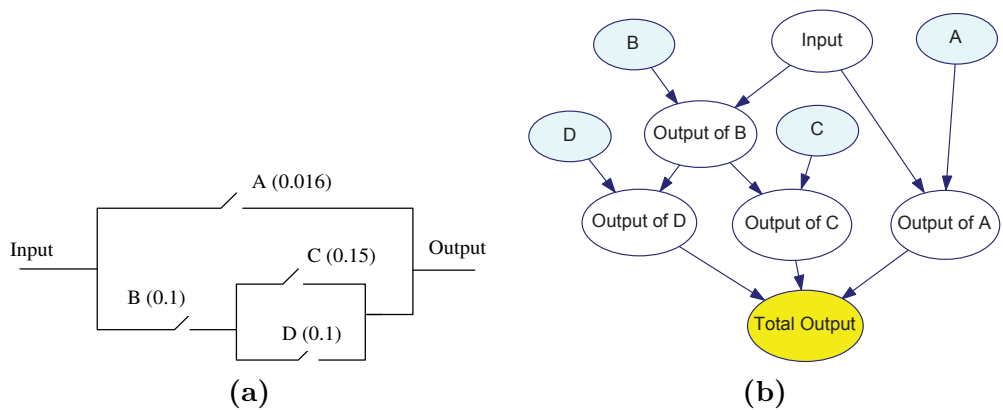

Figure 1: **(a)** An electric circuit and **(b)** its corresponding diagnostic Bayesian network

## 3.2 Explanation in Two Benchmark Models

This section uses a couple of examples to illustrate how some of the methods reviewed in the last section work in practice.

### 3.2.1 Circuit

We first consider the electric circuit in Figure 1(a) (Poole & Provan, 1991). Gates $A, B, C$, and $D$ are *defective* if they are closed. The prior probabilities that the gates are defective are $0.016, 0.1, 0.15$, and $0.1$ respectively. We also assume that the connections between the gates may fail with some small probabilities. The circuit can be modeled with the diagnostic Bayesian network shown in Figure 1(b). Nodes $A, B, C$, and $D$ correspond to the gates in the circuit and have two states each: "defective" and "ok". The others are input or output nodes with two states each as well: "current" or "noCurr". The probabilities that the output nodes of $A, B, C$, and $D$ are in the state "current" given their parent nodes are parameterized as follows.

$$P(Output\ of\ B = current|B = defective, Input = current) = 0.99;$$
$$P(Output\ of\ A = current|A = defective, Input = current) = 0.999;$$
$$P(Output\ of\ C = current|C = defective, Output\ of\ B = current) = 0.985;$$
$$P(Output\ of\ D = current|D = defective, Output\ of\ B = current) = 0.995.$$

Otherwise if no parent is in the state "current", the output nodes are in the state "noCurr" with probability 1.0. Finally, the conditional probability table of "Total Output" is a *noisy-or gate* (Pearl, 1988), which means each parent node that is in the state "current" causes "Total Output" to be in the state "current" independently from the other parents. If no parent node is in the state "current", the probability that "Total Output" is in the state "current" is 0.0. The individual effect of the parent nodes on "Total Output" is parameterized as follows.

$$P(Total\ Output = current|Output\ of\ A = current) = 0.9;$$

$$P(\text{Total Output} = \text{current}|\text{Output of } C = \text{current}) = 0.99;$$
$$P(\text{Total Output} = \text{current}|\text{Output of } D = \text{current}) = 0.995.$$

Suppose we observe that electric current flows through the circuit, which means that nodes *Input* and *Total Output* are both in the state "current". Nodes $A, B, C$, and $D$ are the logical choices of target variables in this model. The task is to find the best explanatory hypothesis to explain the observation of electric current in the circuit. Domain knowledge suggests that there are three basic scenarios that most likely lead to the observation: (1) $A$ is defective; (2) $B$ and $C$ are defective; and (3) $B$ and $D$ are defective.

Given the observation of electric current, the posterior probabilities of $A, B, C$, and $D$ being defective are $0.391, 0.649, 0.446$, and $0.301$ respectively. Therefore, the explanation $(\neg B)$ is the best single-fault explanation, where $\neg B$ means that $B$ is defective. However, $B$ alone does not fully explain the evidence; $C$ or $D$ has to be involved. Actually, if we do not restrict ourselves to the defective states, $(D)$ is the best singleton explanation with probability $0.699$, but it is clearly not a useful explanation for the evidence.

MAP finds $(A, \neg B, \neg C, D)$ as the best explanation. Given that $B$ and $C$ are defective, it is arguable that $A$ and $D$ being okay is irrelevant in explaining the evidence. But MAP has no intrinsic capability to indicate which part of its explanation is more important. Since MPE assumes that all unobserved variables are the target variables, its explanation has even more redundancy.

The pre-pruning techniques are unable to prune any target variable for this model because there is no context-specific independence between the target variables given the evidence. The method of simplifying K-MPE solutions requires the intermediate output nodes to be included as explanatory variables. Since it is typically only necessary to consider the target variables, we adapt their method slightly to simplify the top K MAP solutions instead, which we refer to as the *K-MAP simplification* method hereafter. The best explanation found by this method is $(\neg B, \neg D)$. It is a good explanation, although we will argue later that $(\neg B, \neg C)$ is a better explanation.

The methods based on the likelihood of evidence will overfit and choose $(\neg A, \neg B, \neg C, \neg D)$ as the best explanation, because the probability of the evidence given that all the target variables are defective is almost 1.0.

Both explanation tree methods find $(\neg A)$ as the best explanation. $(\neg A)$ is a good explanation, but again $(\neg B, \neg C)$ is a better explanation as we argue later.

### 3.2.2 Vacation

Consider another example (Shimony, 1993). Mr. Smith is considering taking strenuous hiking trips. His decision to go hiking or not depends on his health status. If he is healthy, he will go hiking; otherwise, he would rather stay home. Mr. Smith is subject to different risks of dying depending on his health status and on where he spends the vacation. The relations between the variables are best represented using the Bayesian network in Figure 2. The conditional probabilities of the model are parameterized as follows:

$$P(\text{healthy}) = 0.8;$$
$$P(\text{home}|\neg \text{healthy}) = 0.8;$$
$$P(\text{home}|\text{healthy}) = 0.1;$$

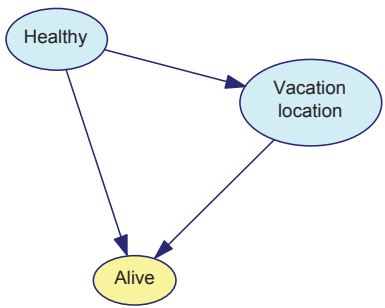

Figure 2: A Bayesian network for the vacation problem.

$$P(alive|healthy, Vacation\ location = *) = 0.99;$$
$$P(alive|\neg healthy, home) = 0.9;$$
$$P(alive|\neg healthy, hiking) = 0.1,$$

where "*" means that the value does not matter. There are totally 100 similar hiking trails for Mr. Smith to choose from. We can model the 100 hiking trips either as different states of the variable "Vacation location", or as one state named "hiking". In case that the hiking trails are modeled as different states, the conditional probability given the health status is distributed evenly across these states. Shimony (1993) showed that the modeling choice of one-state vs. multi-state significantly affects the best explanation of MAP. Given that Mr. Smith is alive after his vacation, the best explanation on the target variable set {Healthy, Vacation location} changes from (healthy, hiking) for the one-state model to (¬healthy, home) for the multi-state model. This is rather undesirable because the explanation should not totally change simply because the model is refined.

We now examine how some other methods are affected by the modeling choice. The explanation tree method finds (hiking) as the explanation for the one-state model and (¬healthy, home) for the multi-state model. These explanations seem counterintuitive. Both the causal explanation tree and K-MAP simplification methods find (healthy) as the explanation for the two models.

What if Mr. Smith died afterwards? This is a more interesting explanation task because the outcome is surprising given the low prior probability of dying in this problem. MAP's explanation changed from (¬healthy, hiking) for the one-state model to (¬healthy, home) for the multi-state model. The explanation tree method finds (hiking) for the one-state model and (home) for the multi-state model, again perplexing explanations. The causal explanation tree method finds (¬healthy, hiking) for the one-state model and (¬healthy, any trip) for the multi-state model; so does the K-MAP simplification method. The causal explanation tree and K-MAP simplification methods are quite robust in the face of the refinement of the model; their explanations also seem plausible. However, since the 100 hiking trips are identical in the multi-state model, any hiking trip can be plugged into the explanation. It means that there are 100 equally good explanations. It is debatable whether the specific hiking trip is a necessary detail that needs to be included.

The above results show that the existing explanation methods in Bayesian networks often generate explanations that are either underspecified or overspecified. They fail to find "just right" explanations that only contain the most relevant target variables.

## 4. Most Relevant Explanation in Bayesian Networks

Users want accurate explanations but do not want to be burdened with unnecessary details. A Bayesian network for a real-world domain may contain many target variables, but typically only a few of the target variables are most relevant in explaining any given evidence. The goal of this research is to develop an explanation method that is able to automatically identify the most relevant target variables in forming its explanation.

We first state several basic assumptions behind this research. First, an explanation typically depends on an agent's epistemic state (Gärdenfors, 1988). We assume that the knowledge encoded in a Bayesian network constitutes the complete epistemic state of the explainer. Second, we assume that the explanandum is specified as the observed states of a set of evidence variables in an explanation task. Third, we assume that the Bayesian network is annotated such that a set of target variables is clearly defined. Under this setting, we are able to focus on the most important issue of how to evaluate and select the best explanation. We believe, however, the proposed methodologies can be easily generalized to more general settings.

### 4.1 What is a Good Explanation?

We consider that a good explanation should have two basic properties: *precise* and *concise*. *Precise* means that the explanation should decrease the surprise value of the explanandum as much as possible. Many other concepts have been used to refer to the same property, including *confirmative* (Carnap, 1948), *sufficient* (Khan, Poupart, & Black, 2009), and *relevant* (Shimony, 1993). We also regard *high explanatory power* (Gärdenfors, 1988) as referring to the preciseness of an explanation. *Concise* means that the explanation should only contain the most relevant variables in explaining the evidence. It is similar to another concept called *minimal* (Khan et al., 2009).

There are attempts to capture both preciseness and conciseness into a single concept, including *consistent* (Pearl, 1988) and *coherent* (Ng & Mooney, 1990). Pearl (1988) argues that an explanation needs to be internally consistent, and that just taking sets of facts that are likely to be true given the evidence may not produce reasonable results. However, putting consistent facts together does not necessarily lead to either preciseness or conciseness. For example, two perfectly correlated facts are consistent, but they are not necessarily relevant in explaining the evidence, and adding both of them to an explanation likely leads to redundancy. Ng and Mooney (1990) define a *coherence metric* that measures how well an explanation ties the various observations together. Their definition also seems likely to lead to simple explanations. However, their definition is based on Horn-clause axioms and cannot be generalized to probabilistic systems easily.

In addition, an explanation method for Bayesian networks should be able to capture the *explaining-away* phenomenon often represented in Bayesian networks. The explaining-away phenomenon refers to the situation in which an effect has multiple causes, and observing the effect and one of the causes reduces the likelihood of the presence of the other causes.

The explaining-away phenomenon can be represented using the collider structure (Pearl, 1988), i.e., the $V$ structure of a single node with multiple parents. It is desirable to capture this phenomenon in order to find explanations that are both precise and concise.

## 4.2 Definition of Explanation of Evidence

We note that the definition of an explanation as a full instantiation of the target variables used in MAP or MPE is quite restrictive. It fundamentally limits the capability of these methods to find concise explanations. In our approach we define *explanation* as follows.

**Definition 1** *Given a set of target variables* $\mathbf{M}$ *in a Bayesian network and partial evidence* $\mathbf{e}$ *on the remaining variables, an* explanation *for the evidence is a joint instantiation* $\mathbf{x}$ *of a non-empty subset* $\mathbf{X}$ *of the target variables, i.e.,* $\emptyset \subset \mathbf{X} \subseteq \mathbf{M}$.

The definition allows an explanation to be any partial instantiation of the target variables. Therefore, it provides an explanation method the freedom to choose which target variables to include.

One key difference between our definition and those of many existing methods is that the existing definitions often have a built-in relevance measure to be optimized, while our definition treats any partial instantiation of the target variables as an explanation. We believe that deciding a relevance measure is a separate issue from defining explanation. The separation of the two issues allows us to not only compare the quality of different explanations but also generalize our method to find multiple top explanations.

Note that we disallow disjunctives in our definition. We agree with Halpern and Pearl (2005) that allowing disjunctive explanations causes both technical and philosophical problems. For example in an explaining-away situation, an effect may have multiple potential causes. Let the number of causes be $n$. Each of the causes can be a good explanation itself. If we allow disjunctives of the causes, there are totally $2^n$ disjunctives. It is really difficult to consider all these disjunctives. Philosophically, if one of the causes is present, any disjunctive that includes that cause as a part is true as well. It is unclear which disjunctive we should choose as the explanation. Besides, a disjunctive of multiple causes seems equivalent to claiming that each cause is a potential explanation. Separating the causes into individual explanations allows comparing the explanations. It is unclear what benefits it has to allow disjunctives.

## 4.3 Relevance Measures

We need a relevance measure to evaluate the quality of an explanation. The measure should be able to not only evaluate the explanatory power of an explanation but also favor more concise explanations so that only the most relevant variables are included in an explanation. Since we are dealing with probabilistic expert systems, the measure should be based on the probabilistic relations between the explanation and the explanandum (Chajewska & Halpern, 1997). It is also desirable if the probabilistic relation can be summarized into a single number (Suermondt, 1992). Next we discuss several popular relevance measures in order to motivate our own choice.

One commonly used measure is the *probability* of an explanation given the evidence, as used in MAP and MPE to find the most likely configuration of a set of target variables. By relying on the posterior probability, however, an explanation may contain independent or marginally relevant events that have a high probability. Methods that use probability as the relevance measure do not have the intrinsic capability to prune the less relevant facts. Some may argue that those variables should not be selected as target variables for explanation in the first place. But that requires the important task of selecting the most relevant variables to rest on the shoulders of users. For an explanation method to be effective, it should perform for the user the task of extracting the most essential knowledge and be as simple as possible (Druzdzel, 1996). Moreover, we have shown earlier that the probability measure is quite sensitive to the modeling choices; simply refining a model can dramatically change the best explanation.

Another commonly used measure is the *likelihood of the evidence* and its variations. The likelihood measure has an undesirable property called *irrelevant conjunction* (Chajewska & Halpern, 1997; Rosenkrantz, 1994), that is, adding an irrelevant fact to a valid explanation does not change its likelihood. This property limits the capability of any method based on this measure to find concise explanations. Another common drawback of the probability and likelihood measures is that they focus on measuring the preciseness of an explanation; they have no intrinsic mechanism to achieve conciseness in an explanation. A measure that can achieve both preciseness and conciseness at the same time is highly desirable.

According to Salmon (1984), in order to construct a satisfactory statistical explanation, it is necessary to factor in both prior and posterior probabilities of either the explanandum or the explanation. An explanation should result from the comparison between the prior and posterior probabilities. This comparative view of explanation generates a set of possibilities. One form of comparison is the difference between the prior and posterior probabilities. However, it is inappropriate as a relevance measure according to Good (1985). Consider an increase of probability from 1/2 to 3/4, and from 3/4 to 1. In both cases the difference in the probabilities is 1/4, but the degree of relevance is entirely different.

Another possibility is the *belief update ratio*, which we define as follows.

**Definition 2** *Assuming $P(x) \neq 0$, the* belief update ratio *of x given e, $r(x;e)$, is defined as*

$$r(x;e) \equiv \frac{P(x|e)}{P(x)} \ .$$ (1)

A trivial mathematical derivation shows that the following is also true.

$$r(x;e) = \frac{P(e|x)}{P(e)}.$$ (2)

Therefore, the belief update ratio is equivalent to the ratio of the posterior and prior probabilities of the explanandum given the explanation. It is clear that both ratios are proportional to the likelihood measure $P(e|x)$ up to a constant. They therefore share the same drawbacks as the likelihood measure.

Suermondt (1992) uses the *cross entropy* between the prior and posterior probability distributions of a target variable to measure the influence of an evidence variable on the target variable. Cross entropy assigns a large penalty for incorrect statements of certainty or near-certainty. However, the measure was used to select the most influential evidence variables, not to select specific states of the target variables as an explanation. Furthermore, it is pointed out that cross-entropy is not an ideal measure because it is not symmetric; it is necessary to keep track of which distribution represents the origin for comparison at all times.

In a 1935 paper (Jeffreys, 1935) and in the book *Theory of Probability* (Jeffreys, 1961), a measure called *Bayes factor* was proposed for quantifying the evidence in favor of a scientific theory. *Bayes factor* is the ratio between the likelihoods of a hypothesis and an alternative hypothesis. If the alternative hypothesis is also a simple statistical hypothesis, the measure is simply the *likelihood ratio*. In other cases when either hypothesis has unknown parameters, the Bayes factor still has the form of a likelihood ratio, but the likelihoods are obtained by integrating over the parameter space (Kass & Raftery, 1995). The logarithm of the Bayes factor was defined as the *weight of evidence* independently by Good (1950) and by Minsky and Selfridge (1961). Similar to cross-entropy, Bayes factor (or the weight of evidence) also assigns large values to the probabilities close to certainty. The drawback of Bayes factor, however, is that it is difficult to use this measure to compare more than two hypotheses (Suermondt, 1992); it is necessary to do pairwise comparisons between multiple hypotheses.

### 4.4 Generalized Bayes Factor

With a slight modification, we can generalize the Bayes factor to compare multiple hypotheses. We define the *generalized Bayes factor* (GBF) in the following.

**Definition 3** *The* generalized Bayes factor *(GBF) of an explanation* $\mathbf{x}$ *for given evidence* $\mathbf{e}$ *is defined as*

$$GBF(\mathbf{x}; \mathbf{e}) \equiv \frac{P(\mathbf{e}|\mathbf{x})}{P(\mathbf{e}|\overline{\mathbf{x}})}, \tag{3}$$

*where* $\overline{\mathbf{x}}$ *denotes the set of all alternative hypotheses of* $\mathbf{x}$.

There is only one hypothesis in $\overline{\mathbf{x}}$ if $\mathbf{X}$ contains a single binary variable. Otherwise, $\overline{\mathbf{x}}$ catches all the alternative hypotheses of $\mathbf{x}$. This *catch-all* form of Bayes factor was previously introduced by Fitelson (2001). In the next few sections, we show that GBF has several desirable theoretical properties that enable it to automatically identify the most relevant variables in finding an explanation. These properties are not possessed by the simple form of Bayes factor and prompt us to give GBF the new name to emphasize its importance.

Note that we do not really need to compute $P(\mathbf{e}|\overline{\mathbf{x}})$ directly in calculating $GBF(\mathbf{x}; \mathbf{e})$. A trivial mathematical derivation shows that

$$GBF(\mathbf{x}; \mathbf{e}) = \frac{P(\mathbf{x}|\mathbf{e})(1 - P(\mathbf{x}))}{P(\mathbf{x})(1 - P(\mathbf{x}|\mathbf{e}))}. \tag{4}$$

Therefore, GBF is no longer a measure for comparing different hypotheses, but simply a measure that compares the posterior and prior probabilities of a single hypothesis. GBF is hence able to overcome the drawback of Bayes factor in having to do pairwise comparisons between multiple hypotheses.

Using $\overline{\mathbf{x}}$ in the definition of GBF to catch all the alternative hypotheses enables GBF to capture a property that we call *symmetry in explanatory power*. Consider two complementary simple hypotheses $H$ and $\neg H$ in the following two distinct cases. In the first case, $P(H)$ increases from 0.7 to 0.8. The increase goes hand in hand with the decrease of $P(\neg H)$ from 0.3 to 0.2. In the second case, $P(H)$ increases from 0.2 to 0.3, which also goes hand in hand with the decrease of $P(\neg H)$ from 0.8 to 0.7. The two cases are completely symmetric to each other, which indicates that the two $H$s should have the same amount of explanatory power. GBF assigns the same score to $H$ in both cases. But many other relevance measures such as probability and the likelihood measure assign them totally different scores.

Besides Equation 4, GBF can be expressed in other forms, one of which is in the following.

$$GBF(\mathbf{x}; \mathbf{e}) \equiv \frac{P(\mathbf{x}|\mathbf{e})/P(\overline{\mathbf{x}}|\mathbf{e})}{P(\mathbf{x})/P(\overline{\mathbf{x}})}. \tag{5}$$

Essentially, GBF is equal to the ratio between the posterior odds ratio of $\mathbf{x}$ given $\mathbf{e}$ and the prior odds ratio of $\mathbf{x}$ (Good, 1985). Therefore, GBF measures the degree of change in the odds ratio of an explanation.

GBF can also be calculated as the ratio between the belief update ratios of $\mathbf{x}$ and alternative explanations $\overline{\mathbf{x}}$ given $\mathbf{e}$, i.e.,

$$GBF(\mathbf{x}; \mathbf{e}) = \frac{P(\mathbf{x}|\mathbf{e})/P(\mathbf{x})}{P(\overline{\mathbf{x}}|\mathbf{e})/P(\overline{\mathbf{x}})}. \tag{6}$$

When there are multiple pieces of evidence, GBF can also be calculated using a *chain rule* similar to that of a joint probability distribution of multiple variables. We first define *conditional Bayes factor* as follows.

**Definition 4** *The* conditional Bayes factor *(CBF) of explanation* $\mathbf{y}$ *for given evidence* $\mathbf{e}$ *conditioned on explanation* $\mathbf{x}$ *is defined as*

$$GBF(\mathbf{y}; \mathbf{e}|\mathbf{x}) \equiv \frac{P(\mathbf{e}|\mathbf{y}, \mathbf{x})}{P(\mathbf{e}|\overline{\mathbf{y}}, \mathbf{x})}. \tag{7}$$

Then, it is easy to show that the following chain rule for calculating the GBF of an explanation given a set of evidence is true.

$$GBF(\mathbf{x}; e_1, e_2, ..., e_n) = GBF(\mathbf{x}; e_1) \prod_{i=2}^{n} GBF(\mathbf{x}; e_i|e_1, e_2, ..., e_{i-1}). \tag{8}$$

The chain rule is especially useful when the multiple pieces of evidence are obtained incrementally.

### 4.4.1 Handling Extreme Values

GBF assigns much more weight to probabilities in ranges close to 0 and 1. Such weighting is consistent with the decision-theoretic interpretation of a subjective probability of 0 and 1 (Suermondt, 1992). The belief that the probability of an event is equal to 1 means that one is absolutely sure that the event will occur. One probably would bet anything, including her own life, on the occurrence of the event. It is a rather strong statement. Therefore, any increase in probability from less than one to one or decrease from non-zero to zero is extremely significant, no matter how small the change is.

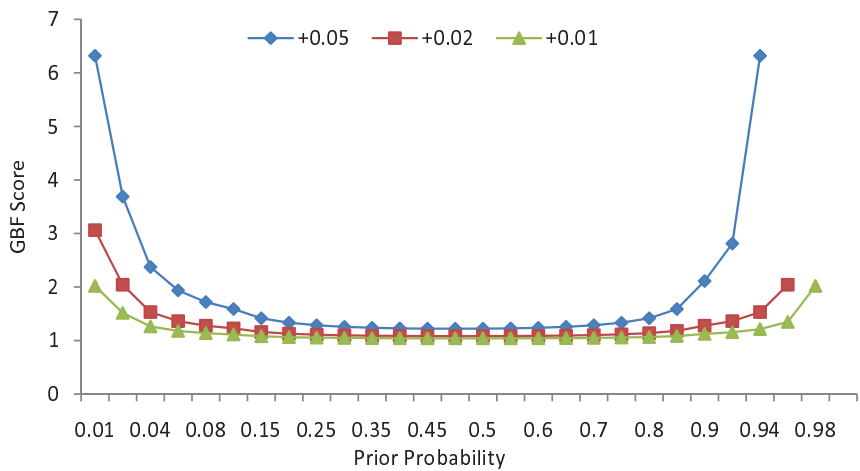

Figure 3: The GBF as a function of the prior probability given a fixed increase in the posterior probability from the prior. The different curves correspond to different probability increases. For example, "+0.01" means the difference between the posterior and prior probabilities is 0.01.

The following are several special cases for computing GBF in the face of extreme probabilities.

If $P(\mathbf{x}) = 0.0$, $P(\mathbf{x}|\mathbf{e})$ must be equal to zero as well, which yields a ratio between two zeros. As is commonly done, we define the ratio of two zeros as zero. Intuitively, an impossible explanation is never useful for explaining any evidence.

If $P(\mathbf{x}) = 1.0$ and $P(\mathbf{x}|\mathbf{e}) = 1.0$, we again have a ratio between two zeros. Since the explanation is true no matter what the evidence is, it is not useful for explaining the evidence either. This case can actually be used as a counterexample for using probability as a relevance measure, because the fact that an explanation has a high posterior probability may simply be due to its high prior probability.

If $P(\mathbf{x}) < 1.0$ and $P(\mathbf{x}|\mathbf{e}) = 1.0$, the GBF score is equal to infinity. The fact that an explanation initially has uncertainty but becomes certainly true after observing the evidence warrants the explanation to have a large GBF score.

### 4.4.2 Monotonicity of GBF

We study the monotonicity of GBF with regard to two other relevance measures. The first measure is the *difference between the posterior and prior probabilities*. It is commonly believed that the same amount of difference in probability in ranges close to zero or one is much more significant than in other ranges. A prominent measure that is able to capture the belief is the K-L divergence (Kullback & Leibler, 1951). A measure for explanatory power should also capture this belief in ranking explanations. Figure 3 shows a plot of GBF against the prior probability when the difference between the posterior and prior probabilities is fixed. For example, "+0.01" means the difference between the posterior and prior probabilities is 0.01. The figure clearly shows that GBF assigns much higher scores to probability changes close to zero and one.

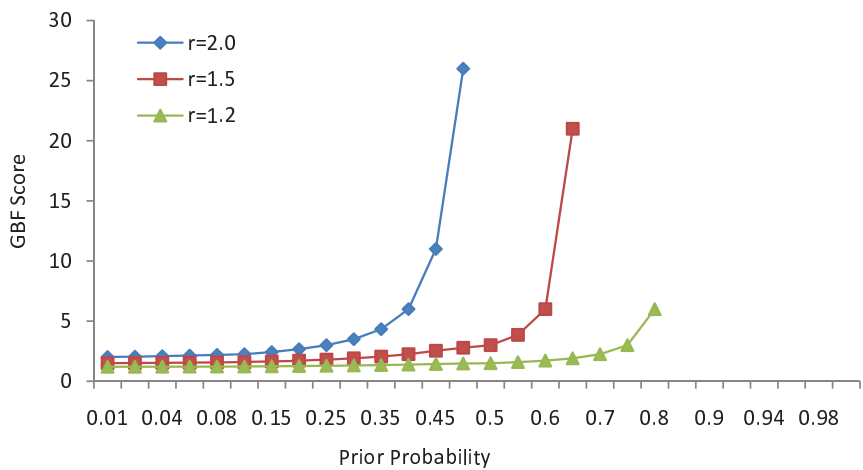

Figure 4: The GBF as a function of the prior probability when the belief update ratio is fixed. The different curves correspond to different belief update ratios.

We also study the monotonicity of GBF with regard to another measure, the belief update ratio defined in Equation 1. Recall that the belief update ratio is proportional to the likelihood measure up to a constant. The likelihood measure, typically together with a penalty term for the complexity of a model, is a popular metric used in model selection. AIC or BIC (Schwartz, 1979) are prominent examples. Next we show that GBF provides more discriminant power than the belief update ratio (hence, also the likelihood measure and the ratio between the posterior and prior probabilities of evidence). We have the following theorem. All the proofs of the theorems and corollaries in this paper can be found in the appendix.

**Theorem 1** *For an explanation* $\mathbf{x}$ *with a fixed belief update ratio* $r(\mathbf{x}; \mathbf{e})$ *greater than 1.0,* $GBF(\mathbf{x}; \mathbf{e})$ *is monotonically increasing as the prior probability* $P(\mathbf{x})$ *increases.*

Figure 4 plots GBF as a function of the prior probability while fixing the belief update ratio. The different curves correspond to different belief update ratios. GBF automatically takes into account the relative magnitude of the probabilities in measuring the quality of an explanation. As the prior probability of an explanation increases, the same ratio of probability increase becomes more and more significant. Therefore, explanations that cannot be distinguished by the likelihood measure can be ranked using GBF. Since lower-dimensional explanations typically have higher probabilities, GBF has the intrinsic capability to penalize more complex explanations. The value pair of <likelihood, prior probability> used by Chajewska and Halpern (1997) is only able to produce partial orderings among the explanations. In some sense, GBF addresses its limitation by integrating the two values into a single value to produce a complete ordering among the explanations. Users do not have the burden to define a complete ordering anymore.

### 4.4.3 Achieving Conciseness in Explanation

This section discusses several theoretical properties of GBF. The key property of GBF is that it is able to weigh the relative importance of multiple variables and only include the most relevant variables in explaining the given evidence. We have the following theorem.

**Theorem 2** *Let the conditional Bayes factor (CBF) of explanation* $\mathbf{y}$ *given explanation* $\mathbf{x}$ *be less than or equal to the inverse of the belief update ratio of the alternative explanations* $\overline{\mathbf{x}}$, *i.e.,*

$$GBF(\mathbf{y};\mathbf{e}|\mathbf{x}) \leq \frac{1}{r(\overline{\mathbf{x}};\mathbf{e})}, \tag{9}$$

*then we have*

$$GBF(\mathbf{x},y;\mathbf{e}) \leq GBF(\mathbf{x};\mathbf{e}). \tag{10}$$

The conditional independence relations in Bayesian networks only provide a hard measure on the relevance of an explanation $\mathbf{y}$ with regard to another explanation $\mathbf{x}$; the answer is either yes or no. In contrast, $GBF(\mathbf{y};\mathbf{e}|\mathbf{x})$ is able to provide a soft measure on the relevance of $\mathbf{y}$ in explaining $\mathbf{e}$ given $\mathbf{x}$. $GBF$ also encodes a decision boundary, the *inverse belief update ratio* of the alternative explanations $\overline{\mathbf{x}}$ given $\mathbf{e}$. The ratio is used to decide how important the remaining variables should be in order to be included in an explanation. If $GBF(\mathbf{y};\mathbf{e}|\mathbf{x})$ is greater than $\frac{1}{r(\overline{\mathbf{x}};\mathbf{e})}$, $\mathbf{y}$ is important enough to be included. Otherwise, $\mathbf{y}$ will be excluded from the explanation. Simply being dependent is not enough for a variable to be included in an explanation.

Theorem 2 has several intuitive and desirable corollaries. The first corollary states that, for any explanation $\mathbf{x}$ with a belief update ratio greater than 1.0, adding an independent variable to the explanation will decrease its GBF score.

**Corollary 1** *Let* $\mathbf{x}$ *be an explanation with* $r(\mathbf{x};\mathbf{e}) > 1.0$, *and* $Y \perp \mathbf{X}, \mathbf{E}$, *then for any state* $y$ *of* $Y$, *we have*

$$GBF(\mathbf{x},y;\mathbf{e}) < GBF(\mathbf{x};\mathbf{e}). \tag{11}$$

Therefore, adding an irrelevant variable will dilute the explanatory power of an existing explanation. GBF is able to automatically prune such variables from its explanation.

Note that we focus on explanations with a belief update ratio greater than 1.0. An explanation whose probability does not change or even decreases given the evidence does not seem to be able to relieve the "cognitive dissonance" between the explanandum and the rest of our beliefs. According to Chajewska and Halpern (1997), the fact that a potential explanation is equally or less likely a posteriori than a priori should cause some suspicion. In the philosophical literature, it is also often required that the posterior probability of the explanandum to be at least greater than its unconditional probability, so that learning the explanation increases the probability of the explanandum (Gärdenfors, 1988; Carnap, 1948).

Corollary 1 requires that variable $Y$ to be independent from both $\mathbf{X}$ and $\mathbf{E}$. The assumption is rather strong. The following corollary shows that the same result holds if $Y$ is conditionally independent from $\mathbf{E}$ given $\mathbf{x}$.

**Corollary 2** *Let* **x** *be an explanation with* $r(\mathbf{x}; \mathbf{e}) > 1.0$, *and* $Y \perp \mathbf{E}|\mathbf{x}$, *then for any state* $y$ *of* $Y$, *we have*

$$GBF(\mathbf{x}, y; \mathbf{e}) < GBF(\mathbf{x}; \mathbf{e}). \tag{12}$$

Corollary 2 is a more general result than Corollary 1 and captures the intuition that conditionally independent variables add no additional information to an explanation in explaining the evidence. Note that these properties are all relative to an existing explanation. It is possible that a variable is independent from the evidence given one explanation, but becomes dependent on the evidence given another explanation. Therefore, selecting variables one by one greedily does not guarantee to find the explanation with the highest GBF.

The above results can be further relaxed to accommodate cases in which the posterior probability of $y$ given $\mathbf{e}$ is smaller than its prior conditioned on $\mathbf{x}$, i.e.,

**Corollary 3** *Let* **x** *be an explanation with* $r(\mathbf{x}; \mathbf{e}) > 1.0$, *and* $y$ *be a state of a variable* $Y$ *such that* $P(y|\mathbf{x}, \mathbf{e}) < P(y|\mathbf{x})$, *then we have*

$$GBF(\mathbf{x}, y; \mathbf{e}) < GBF(\mathbf{x}; \mathbf{e}). \tag{13}$$

This is again an intuitive result; the state of a variable whose posterior probability decreases given the evidence should not be part of an explanation for the evidence.

We applied GBF to rank the candidate explanations in the circuit example introduced in Section 3. The following shows a partial ranking of the explanations with the highest GBF scores.

$$GBF(\neg B, \neg C; \mathbf{e}) > GBF(\neg B, \neg C, A; \mathbf{e}), GBF(\neg B, \neg C, D; \mathbf{e}) > GBF(\neg B, \neg C, A, D; \mathbf{e})$$

$(\neg B, \neg C)$ not only has the highest GBF score but also is more concise than the other explanations, which indicates that variables $A$ and $D$ are not very relevant in explaining the evidence once $(\neg B, \neg C)$ is observed. The results indicate that GBF has the intrinsic capability to penalize higher-dimensional explanations and prune less relevant variables, which match the theoretical properties well.

## 4.5 Most Relevant Explanation

The theoretical properties presented in the previous section show that GBF is a plausible relevance measure for the explanatory power of an explanation. In particular, they show that GBF is able to automatically identify the most relevant target variables in finding an explanation. We hence propose a method called *Most Relevant Explanation* (MRE) which relies on GBF in finding explanations for given evidence in Bayesian networks.

**Definition 5** *Let* **M** *be a set of target variables, and* **e** *be the partial evidence on the remaining variables in a Bayesian network.* Most Relevant Explanation *is the problem of finding an explanation* **x** *for* **e** *that has the maximum* generalized Bayes factor *score* $GBF(\mathbf{x}; \mathbf{e})$, *i.e.,*

$$MRE(\mathbf{M}; \mathbf{e}) \equiv \arg\max_{\mathbf{x}, \emptyset \subset \mathbf{X} \subseteq \mathbf{M}} GBF(\mathbf{x}; \mathbf{e}) \, . \tag{14}$$

Although MRE is general enough to be applied to any probabilistic distribution model, MRE's properties make it especially suitable for Bayesian networks. Bayesian networks were invented to model the conditional independence relations between the random variables of a domain so that we not only obtain a concise representation of the domain but also have efficient algorithms for reasoning about the relations between the variables. The concise representation of a Bayesian network is also beneficial to MRE in the following ways.

First, MRE can utilize the conditional independence relations modeled by a Bayesian network to find explanations more efficiently. For example, Corollaries 1 and 2 can be used to prune independent and conditionally independent target variables so that not all of the partial instantiations of the target variables need to be considered during the search for a solution. Such independence relations can be identified through the graphical structure of a Bayesian network and may significantly improve the efficiency of the search for an explanation.

Second, similar to the chain rule of Bayesian networks, the chain rule of GBF in Equation 7 can also be simplified using the conditional independence relations. When computing $GBF(\mathbf{x}; e_i | e_1, e_2, ..., e_{i-1})$, we may only need to condition on a subset of the evidence in $(e_1, e_2, ..., e_{i-1})$. Conditioning on fewer evidence variables allows the reasoning algorithms to involve a smaller part of a Bayesian network in computing the GBF score (Lin & Druzdzel, 1998). One extreme case is that, when all the evidence variables are independent given an explanation, the GBF of the explanation given all the evidence is simply the product of the individual GBFs given each individual piece of evidence.

$$GBF(\mathbf{x}; e_1, e_2, ..., e_n) = \prod_{i=1}^{n} GBF(\mathbf{x}; e_i). \tag{15}$$

Finally, MRE is able to capture well the unique *explaining-away* phenomenon that is often represented in Bayesian networks. Wellman and Henrion (1993) characterized *explaining away* as the *negative influence* between two parents induced by an observation on the child in a Bayesian network. Let $A$ and $B$ be predecessors of $C$ in a graphical model G. $C$ is observed to be equal to $c$. Let $x$ denote an assignment to $C$'s other predecessors, and $y$ denote an assignment to $B$'s predecessors. The *negative influence* between $A$ and $B$ conditioned on $C = c$ means

$$P(B|A, c, x, y) \le P(B|c, x, y) \le P(B|\neg A, c, x, y). \tag{16}$$

MRE captures the explaining-away phenomenon well with CBF. CBF provides a measure on how relevant some new variables are in explaining the evidence conditioned on an existing explanation. In the explaining-away situation, if one of the causes is already present in an explanation, the other causes typically do not receive high CBFs. In fact, CBF can capture Equation 16 in an equivalent way as shown in the following theorem.

**Theorem 3** *Let $A$ and $B$ be predecessors of $C$ in a Bayesian network. $C$ is observed to be equal to $c$. Let $\mathbf{x}$ denote an assignment to $C$'s other predecessors, and $\mathbf{y}$ denote an assignment to $B$'s predecessors, then we have*

$$P(B|A, c, \mathbf{x}, \mathbf{y}) \le P(B|c, \mathbf{x}, \mathbf{y}) \le P(B|\neg A, c, \mathbf{x}, \mathbf{y}) \tag{17}$$

$$\Leftrightarrow \quad GBF(B; c|A, \mathbf{x}, \mathbf{y}) \le GBF(B; c|\mathbf{x}, \mathbf{y}) \le GBF(B; c|\neg A, \mathbf{x}, \mathbf{y}) \quad . \tag{18}$$

| Bayes factor | Strength of evidence |
|---|---|
| <1 | Negative |
| 1 to 3 | Barely worth mentioning |
| 3 to 10 | Substantial |
| 10 to 30 | Strong |
| 30 to 100 | Very strong |
| >100 | Decisive |

Table 1: Bayes factor.

Again consider the circuit example, $(\neg B, \neg C)$ and $(\neg A)$ are both good explanations for the evidence by themselves; the GBF scores of $(\neg B, \neg C)$ and $(\neg A)$ given only $\mathbf{e}$ are 42.62 and 39.44 respectively. Jeffreys (1961) recommends using Table 1 as a guidance in determining the significance of a Bayes factor value. If we use the same table to judge GBFs, $(\neg B, \neg C)$ and $(\neg A)$ are both very strong explanations. However, when $(\neg B, \neg C)$ is already observed, $GBF(\neg A; \mathbf{e}|\neg B, \neg C)$ is only equal to 1.03, which is barely worth mentioning. The results indicate that GBF is able to capture the explaining-away phenomenon for the circuit example.

### 4.6 K-MRE

In many decision making problems, decision makers typically would like multiple competing options to choose from. Outputting only the best solution is hence not the best practice. This is especially true when there are multiple top solutions that are almost equally good. In that case, we want to know not only which solution is the best, but also how much better it is than the other solutions. If the difference between the first and second best solutions is large, it gives decision makers more confidence in the quality of the best solution. Moreover, finding multiple top solutions can be used as a sensitivity analysis method to provide insight on how sensitive the best explanations are to the changes in the model parameters.

A naive approach to finding the top K MRE solutions is to select the explanations with the highest GBF scores. However, this strategy may find explanations that are supersets of other top explanations. Once again consider the circuit example. Table 2 lists the explanations with the highest GBF scores. Simply relying on the scores will produce the following rather similar top four explanations: $(\neg B, \neg C)$, $(A, \neg B, \neg C)$, $(\neg B, \neg C, D)$, and $(A, \neg B, \neg C, D)$. All these explanations essentially cover the same basic scenario in which both $B$ and $C$ are defective. We have to search further down the list before finding the other two basic scenarios: $A$ is defective, and $B$ and $D$ are both defective.

It is often critical to achieve *diversity* when the goal is to find multiple solutions. For example, it is argued that diversity is important in recommender systems (Smyth & Mc-Clave, 2001). A rather similar set of recommendations will not give users a useful set of alternatives to choose from. We believe that the same holds in finding explanations. In order to find a set of top explanations that are more diverse and representative, we define two dominance relations among the candidate solutions of MRE. The first relation is *strong dominance.*

| Explanations | GBF |
|---|---|
| $(\neg\mathbf{B}, \neg\mathbf{C})$ | 42.62 |
| $(A, \neg B, \neg C)$ | 42.15 |
| $(\neg B, \neg C, D)$ | 39.93 |
| $(A, \neg B, \neg C, D)$ | 39.56 |
| $(\neg\mathbf{A})$ | 39.44 |
| $(\neg A, B)$ | 36.98 |
| $(\neg A, C)$ | 35.99 |
| $(\neg\mathbf{B}, \neg\mathbf{D})$ | 35.88 |

Table 2: The explanations with the highest GBF scores for the Circuit network. The explanations in boldface are the top *minimal* explanations.

**Definition 6** *An explanation* $\mathbf{x}$ strongly dominates *another explanation* $\mathbf{y}$ *if and only if* $\mathbf{x} \subset \mathbf{y}$ *and* $GBF(\mathbf{x}) \geq GBF(\mathbf{y})$.

If $\mathbf{x}$ strongly dominates $\mathbf{y}$, $\mathbf{x}$ is clearly a better explanation than $\mathbf{y}$, because it not only has a higher or equal explanatory score but also is more concise. We only need to include $\mathbf{x}$ in the top explanation set. The second relation is *weak dominance*.

**Definition 7** *An explanation* $\mathbf{x}$ weakly dominates *another explanation* $\mathbf{y}$ *if and only if* $\mathbf{x} \supset \mathbf{y}$ *and* $GBF(\mathbf{x}) > GBF(\mathbf{y})$.

In this case, $\mathbf{x}$ has a larger $GBF$ score than $\mathbf{y}$, but $\mathbf{y}$ is more concise. It is possible that we can include them both and let the decision maker to decide whether she prefers conciseness or a higher score. However, we believe that we only need to include $\mathbf{x}$, because its higher GBF score indicates that the extra variables in $\mathbf{X}$ but not in $\mathbf{Y}$ are important in explaining the given evidence.

Based on the two kinds of dominance relations, we define the concept *minimal*.

**Definition 8** *An explanation is* minimal *if it is neither strongly nor weakly dominated by any other explanation.*

A new K-MRE approach can now be defined such that only the *minimal* explanations are included in the top set. For the circuit network, since $(A, \neg B, \neg C)$, $(\neg B, \neg C, D)$, and $(A, \neg B, \neg C, D)$ are strongly dominated by $(\neg B, \neg C)$, we only need to consider $(\neg B, \neg C)$ among them. Similarly, $(\neg A, B)$ and $(\neg A, C)$ are strongly dominated by $(\neg A)$, so only $(\neg A)$ is included in the top set. Finally, we get the set of top explanations shown in boldface in Table 2. It is clearly more diverse and representative than the original set that contains the dominated explanations.

The dominance relations defined here are not restricted to GBF; they are also applicable to other relevance measures. For example, they can potentially help the methods based on the likelihood measure to find more concise explanations.

## 5. Case Studies

We tested MRE on the explanation tasks of a set of benchmark Bayesian networks from the literature, including Circuit (Poole & Provan, 1991), Vacation (Shimony, 1993), Academe (Flores et al., 2005), Asia (Lauritzen & Spiegelhalter, 1988), and Circuit2 (Darwiche, 2009). These networks were annotated such that the variables are classified into three categories: *target*, *observation*, and *auxiliary*. A target node, also named fault, usually represents a diagnostic interest (e.g., the health status of an engine). An observation node usually represents a symptom (e.g., observing excessive smoking in the engine exhaust), built-in error message (e.g., the status of a power supply), or a test (e.g., measuring the voltage of a battery). A node which is neither target nor observation is classified as an auxiliary node.

We compared K-MRE against several existing explanation methods, including K-MAP (Pearl, 1988), explanation tree (ET) (Flores et al., 2005), causal explanation tree (CET) (Nielsen et al., 2008), and K-MAP simplification (K-SIMP) (de Campos et al., 2001). Several of these methods have tunable parameters. The explanation tree (ET) method has two parameters for controlling the growth of an explanation tree. One parameter is a threshold value for deciding whether a target variable is significant enough to be used to expand a branch of the explanation tree; the variable is used if the average mutual information between the target variable and the other unused target variables conditioned on the current branch is larger than or equal to the threshold. The other parameter is a threshold value on the probability of any branch of the explanation tree. A branch is not expanded further if the probability of the branch is less than the threshold. We set these two parameters to be 0.05 and 0 respectively. A branch in the explanation tree is marked with its posterior probability.

The causal explanation tree (CET) method has only one parameter, a lower-bound threshold on the causal information flow of a variable on the evidence conditioned on the current branch. If the causal information flow is larger than or equal to the threshold, the variable is used to expand the branch further. We set the threshold to be 0.01. Each branch of a causal explanation tree is marked using the log ratio of the posterior and prior probabilities of the evidence given the branch.

The K-MAP simplification (K-SIMP) method also has a threshold value which bounds the reduction in the likelihood of evidence when simplifying an explanation. If deleting a variable only reduces the likelihood of an explanation within a factor bounded by the threshold, the simplification is allowed. We set the threshold value to be 0.05. We keep track of the explanations with the highest likelihood values during the simplification.

All the parameters in these methods were set to allow as much expansion and simplification as possible. But even so, the ET and CET methods may fail to find any significant variable to create even the root of an explanation tree. When that happens, we force these two methods to generate at least the root node by ignoring the thresholds.

### 5.1 Circuit

For the circuit network introduced in Section 3, Table 3 lists the top explanations found by K-MRE, K-MAP, and K-SIMP. We set K to be 3 throughout the case studies. Figure 5 shows the explanation trees found by the ET and CET methods.

| Circuit | Explanations | Scores |
|---------|--------------|--------|
| K-MRE  | $(\neg B, \neg C)$ | 42.62 |
|         | $(\neg A)$ | 39.45 |
|         | $(\neg B, \neg D)$ | 35.88 |
| K-MAP  | $(A, \neg B, \neg C, D)$ | 0.0128 |
|         | $(\neg A, B, C, D)$ | 0.0099 |
|         | $(A, \neg B, C, \neg D)$ | 0.0082 |
| K-SIMP | $(\neg B, \neg D)$ | 0.9818 |
|         | $(\neg B, \neg C)$ | 0.9683 |
|         | $(\neg A)$ | 0.9014 |

Table 3: Top explanations found by K-MRE, K-MAP, and K-SIMP for the Circuit network given the observation of electric current. Note the scores of the methods have the following different meanings: GBF for MRE, probability for MAP, and the likelihood of evidence for K-SIMP.

K-MRE was able to find intuitive explanations for the Circuit network. $(\neg B, \neg C)$ is a better explanation than both $(\neg A)$ and $(\neg B, \neg D)$, because it has a larger posterior probability than the other two explanations (the posterior probabilities are 0.394, 0.391, and 0.266 respectively), while the prior probabilities of the explanations are 0.015, 0.016, and 0.01 respectively.

The explanations found by K-MAP are mostly consistent with what K-MRE found, although the K-MAP solutions are all supersets of the K-MRE solutions. It is not surprising because MAP has to find complete assignments to the target variables. MAP has no ability to indicate which parts of the explanations are most important, which we believe is a fundamental drawback of MAP. Users are now burdened by the task of identifying the most important parts of the explanations.

The K-SIMP method first finds the top K MAP solutions and then simplifies them by deleting variables that do not reduce the likelihood of evidence much, if any. The set of top solutions found by K-SIMP is the same as that of K-MRE. The results indicate that the simplification method helped to prune less relevant target variables for the Circuit network. However, K-SIMP's ranking of the explanations is different. $(\neg B, \neg D)$ is its best explanation. The number of variables of an explanation was also considered as a ranking criterion for the explanations (de Campos et al., 2001), but then $(\neg A)$ becomes the best explanation.

The ET method selects node $A$ as the root of the explanation tree. $A$ is important because whether $A$ is closed or not significantly affects the likelihood of the evidence. However, the ET method selects variable $D$ as the second most important variable, which does not lead to any good explanation. Moreover, there is no easy way to extract the top explanations from the explanation tree. The ET method relies on probability in ranking the explanations. It seems that we should not consider partial paths as the solutions. For example, $(A)$ has a higher probability than $(\neg A)$ but is clearly not a good explanation. If we only consider the full paths from the root to the leaves, $(\neg A)$ is the best explanation.

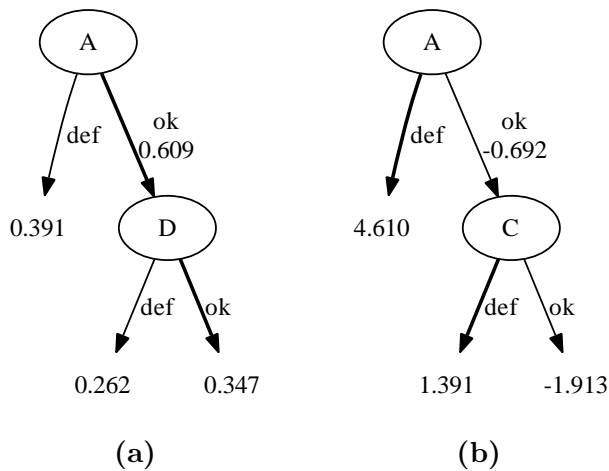

Figure 5: **(a)** Explanation tree and **(b)** causal explanation tree for the Circuit network given the observation of electric current.

Although $(A, D)$ has a probability that is only slightly smaller than $(\neg A)$, it is not a good explanation at all. Moreover, using probability to rank the explanations inevitably makes the threshold value for bounding the probabilities of the branches have a significant effect on the ranking, which we believe is a fundamental flaw of the ET method.

The CET method also selects node $A$ as the root of its explanation tree. However, it selects $C$ as the second most important variable, which does not lead to any good explanation either. Since the branches of a causal explanation tree are marked by the log ratio of the posterior and prior probabilities of the evidence, it makes sense to consider all the partial branches as explanations. The top two explanations according to the CET method are $(\neg A)$ and $(A, \neg C)$, but $(A, \neg C)$ is clearly not a good explanation.

Both explanation tree methods failed to find either $(\neg B, \neg C)$ or $(\neg B, \neg D)$ as a top explanation. The main reason is that they are *greedy* search methods. They may be good at identifying individual variables that are important, but they often fail to identify the compound effect of multiple variables. Even though variable $B$ itself may not have as large an effect as $A$, it forms excellent explanations together with $C$ or $D$. The explanation tree methods both failed to find these variable pairs as they only consider one variable at a time.

### 5.2 Vacation

Two versions of the vacation network were introduced in Section 3. One version models all the possible hiking trips as one state named "hiking", while the other version models the 100 hiking trips as separate but identical states. For both networks, we first consider the scenario in which Mr. Smith is alive after his vacation. Table 4 shows the top explanations by K-MRE, K-MAP, and K-SIMP, and Figure 6 shows the explanation trees by the ET and CET methods.

| Vacation | One-state model | | Multi-state model | |
|---|---|---|---|---|
| | Explanations | Scores | Explanations | Scores |
| K-MRE | $(healthy)$ | 1.3378 | $(healthy)$ | 1.3378 |
| | $(home)$ | 1.0078 | $(any\ trip)$ | 1.0034 |
| K-MAP | $(healthy, hiking)$ | 0.6336 | $(\neg healthy, home)$ | 0.1440 |
| | $(healthy, home)$ | 0.1584 | $(healthy, home)$ | 0.0792 |
| | $(\neg healthy, home)$ | 0.1440 | $(healthy, any\ trip)$ | 0.0071 |
| K-SIMP | $(healthy)$ | 0.9900 | $(healthy)$ | 0.9900 |
| | $(home)$ | 0.9450 | $(home)$ | 0.9300 |

Table 4: Top explanations found by K-MRE, K-MAP, and K-SIMP for the two Vacation networks given that Mr. Smith is alive after his vacation.

K-MRE only found two top explanations for each model because the other explanations all have GBFs less than or equal to 1.0. $(healthy)$ is the best explanation for both models. In fact, the explanation has the same score in both models. Vacation location is treated by K-MRE as irrelevant because the probability of staying alive is the same regardless of where Mr. Smith decides to spend the vacation. Although the second best explanation changed from $(home)$ to $(any\ tip)$, we note that both explanations have GBF scores very close to 1.0 and are barely worth mentioning according to Table 1. Actually the best explanations also have GBFs only slightly over 1.0 and are not interesting explanations either. The reason is that Mr. Smith's being alive after his vacation is not a surprising event; there is no real need for explaining the observation.

K-MAP is extremely sensitive to the modeling choice. Not only did the best explanation change from $(healthy, hiking)$ to $(\neg healthy, home)$, but also the highest probability decreased significantly from 0.6336 to 0.1440. $(\neg healthy, home)$ is ranked third in the one-state model, but became the best explanation in the multi-state model.

Although the K-SIMP method started by simplifying the top three explanations by K-MAP, it ended up with only two explanations for both models. It is because the simplification method resulted in duplicate solutions. Otherwise, the simplification method is quite robust in the face of the modeling choice; $(healthy)$ is the best explanation in both models.

The ET method produced similar trees for both models. However, it selects Vacation location as the most important variable. This is counterintuitive because Vacation location does not even affect the probability of Mr. Smith being alive when he is healthy. This result indicates that the mutual information between the target variables is not a good indicator of their relevance in explaining the evidence. Also, it is again unclear which explanations we should extract from the tree. If we rely on probability in selecting full branches, we will select $(hiking)$ for the one-state model and $(\neg healthy, home)$ for the multi-state model. Neither of them is a good explanation for Mr. Smith's being alive after his vacation.

The causal explanation tree method is also robust in the face of the modeling choice. It selects Healthy as the most important variable. It also finds $(healthy)$ as the best explanation in both models.

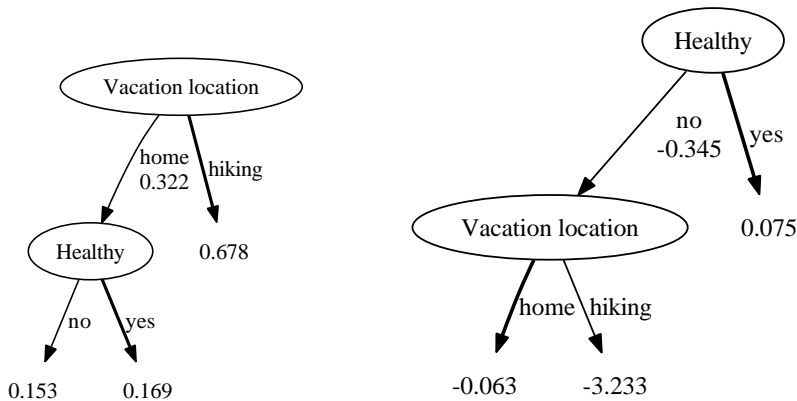

**(a)** ET for the one-state model   **(b)** CET for the one-state model

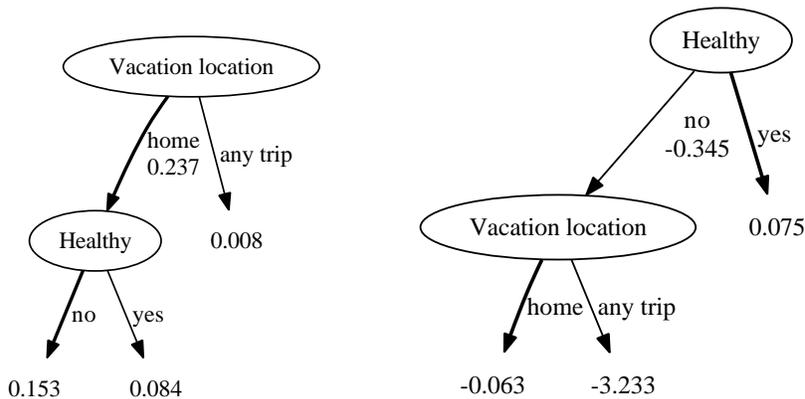

**(c)** ET for the multi-state model   **(d)** CET for the multi-state model

Figure 6: Explanation trees (ET) and causal explanation trees (CET) for the Vacation networks give that Mr. Smith is alive after his vacation.

Next we consider the scenario in which Mr. Smith died after his vacation. Table 5 and Figure 7 show the explanations found by the various methods for this observation.

MRE finds $(\neg healthy, hiking)$ as the best explanation for the one-state model and $(\neg healthy)$ for the multi-state model. It is a rather intuitive result. For the one-state model, hiking is a likely event and significantly increases the chance of Mr. Smith's death if he is unhealthy. For the multi-state model, however, the hiking trails are modeled as individual states, each of which has rather low prior and posterior probabilities. As a result, MRE considered the individual hiking trips as unimportant details and excluded them from the best explanation. Some may argue that each individual hiking trip has the same effect as the hiking state in the one-state model. However, that reasoning ignores the prior probabilities of the explanations and essentially supports the use of the likelihood measure to select explanations. As we already discussed earlier, the likelihood measure has some significant drawbacks.

| Vacation | One-state model | | Multi-state model | |
|---|---|---|---|---|
| | Explanations | Scores | Explanations | Scores |
| K-MRE | $(\neg healthy, hiking)$ | 36.00 | $(\neg healthy)$ | 26.0000 |
| | | | $(home)$ | 1.2310 |
| K-MAP | $(\neg healthy, hiking)$ | 0.0360 | $(\neg healthy, home)$ | 0.0160 |
| | $(\neg healthy, home)$ | 0.0160 | $(healthy, home)$ | 0.0008 |
| | $(healthy, hiking)$ | 0.0064 | $(\neg healthy, any\ trip)$ | 0.0004 |
| K-SIMP | $(\neg healthy, hiking)$ | 0.9000 | $(\neg healthy, any\ trip)$ | 0.9000 |
| | $(\neg healthy)$ | 0.2600 | $(\neg healthy)$ | 0.2600 |
| | $(hiking)$ | 0.0624 | $(home)$ | 0.0700 |

Table 5: Top explanations found by K-MRE, K-MAP, and K-SIMP for the two Vacation networks given that Mr. Smith died after his vacation.

K-MAP is again shown to be sensitive to the modeling choice. The best explanation changed from $(\neg healthy, hiking)$ for the one-state model to $(\neg healthy, home)$ for the multi-state model. $(\neg healthy, home)$ is not a good explanation because staying home reduces the likelihood of Mr. Smith's death after his vacation.

The K-SIMP method selects $(\neg healthy, hiking)$ for the one-state model and $(\neg healthy, any\ trip)$ for the multi-state model. Therefore, there are 100 best explanations with exactly the same score according to this method.

The ET method creates simple trees with only a root for both models. However, the variable chosen by this method is again counterintuitive. Vacation location is not the most important variable for explaining Mr. Smith's death.

The CET method made a good choice in selecting Healthy as the most important variable. Similar to K-SIMP, CET also included the detailed vacation locations as part of its explanations, so it has 100 best explanations with the same score.

## 5.3 Academe

Figure 8 shows the Academe network introduced by Flores et al. (2005) to discuss the explanation tree method. The prior probability distributions of the four target variables Theory, Practice, Extra, and OtherFactors are parameterized as follows.

$$P(Theory) < good, average, bad > \ = \ < 0.4, 0.3, 0.3 >;$$
$$P(Practice) < good, average, bad > \ = \ < 0.6, 0.25, 0.15 >;$$
$$P(Extra) < yes, no > \ = \ < 0.3, 0.7 >;$$
$$P(OtherFactors) < plus, minus > \ = \ < 0.8, 0.2 >;$$

The conditional probabilities of MarkTP, GlobalMark, and FinalMark given their parents are parameterized as follows.

$$P(MarkTP = pass | Theory = bad \lor Practice = bad) \ = \ 0.0;$$

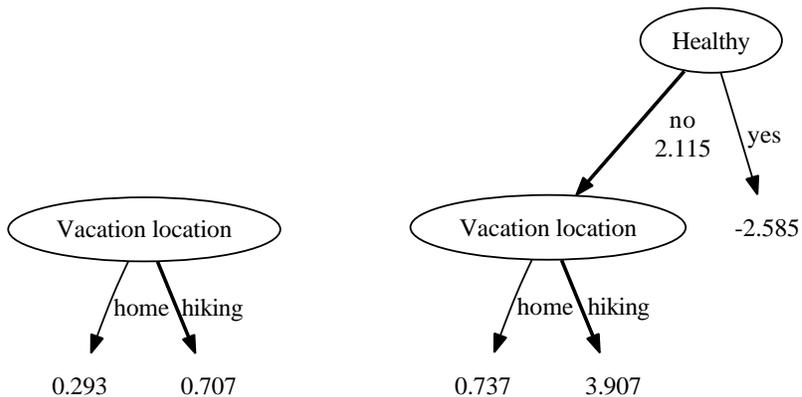

**(a)** ET for the one-state model      **(b)** CET for the one-state model

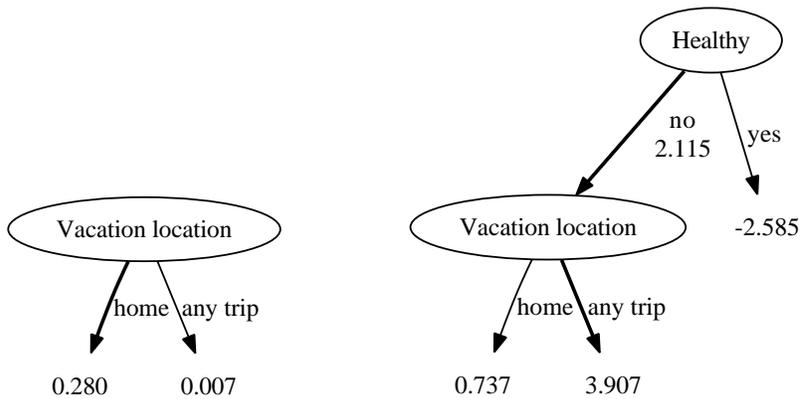

**(c)** ET for the multi-state model      **(d)** CET for the multi-state model

Figure 7: Explanation trees (ET) and causal explanation trees (CET) for the Vacation networks given that Mr. Smith died after his vacation.

$$P(MarkTP = pass|Theory = good, Practice = good) = 1.0;$$
$$P(MarkTP = pass|Theory = good, Practice = average) = 0.85;$$
$$P(MarkTP = pass|Theory = average, Practice = good) = 0.9;$$
$$P(MarkTP = pass|Theory = average, Practice = average) = 0.2;$$
$$P(GlobalMark = pass|MarkTP = pass, Extra = *) = 1.0;$$
$$P(GlobalMark = pass|MarkTP = fail, Extra = yes) = 0.25;$$
$$P(GlobalMark = pass|MarkTP = fail, Extra = no) = 0.0;$$
$$P(FinalMark = pass|GlobalMark = pass, OtherFactors = plus) = 1.0;$$
$$P(FinalMark = pass|GlobalMark = pass, OtherFactors = minus) = 0.7;$$

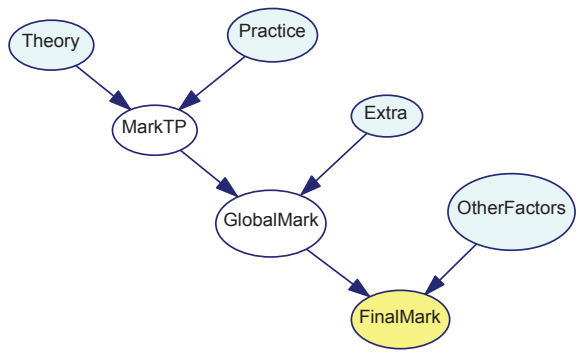

Figure 8: The Academe network.

| Academe | Explanations | Scores |
|---------|-------------|--------|
| K-MRE | (*bad theory*) | 3.0205 |
| | (*bad practice, no extra*) | 2.2986 |
| | (*good theory, bad practice, minus otherFactors*) | 2.0209 |
| K-MAP | (*bad theory, good practice, no extra, plus otherFactors*) | 0.0958 |
| | (*bad theory, average practice, no extra, plus otherFactors*) | 0.0399 |
| | (*average theory, bad practice, no extra, plus otherFactors*) | 0.0399 |
| K-SIMP | (*bad theory, no extra*) | 0.9600 |
| | (*average theory, bad practice*) | 0.7260 |

Table 6: Top explanations found by K-MRE, K-MAP, and K-SIMP for the Academe network given that the FinalMark is fail.

$$P(FinalMark = pass | GlobalMark = fail, OtherFactors = plus) = 0.05;$$
$$P(FinalMark = pass | GlobalMark = fail, OtherFactors = minus) = 0.0.$$

We consider the problem of finding explanations for the observation that FinalMark is fail using the target variable set {*Theory, Practice, Extra, OtherFactors*}. Table 6 and Figure 9 show the explanations found by the various methods for this observation.

K-MRE selects (*bad theory*) as the best explanation for the failing final grade. There are other good explanations that contain bad theory as a part, but (*bad theory*) dominates all those explanations. (*bad practice*) itself is a good explanation, but the combination of bad practice and no extra preparation turns out to be a better explanation.

The top three explanations found by K-MAP all have probabilities smaller than 0.1. In general, it is highly domain-dependent as to which probabilities mean good explanations and which mean bad explanations. In comparison, GBFs and the likelihood of evidence are more consistent across different domains. Note that we are not claiming that the top explanations ranked by probability are necessarily bad; we just believe that probability may not be a good measure for the explanatory power of an explanation.

The K-SIMP method selects (*bad theory, no extra*) as the best explanation. This explanation is a good explanation itself. However, it is not really the explanation with the highest likelihood. For example, the likelihood of a failing grade given (*bad theory*, *bad practice*, *no extra*, *minus otherFactors*) is equal to 1.0, but K-SIMP is limited to the top solutions found by K-MAP. Also, the method again ended up with only two explanations, because the top two MAP solutions are simplified to the same explanation.

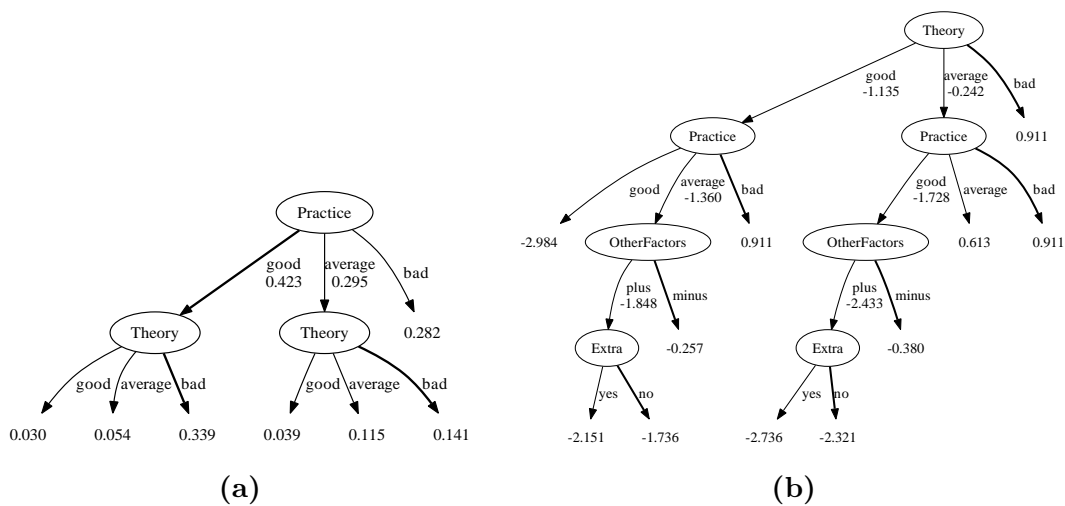

Figure 9: **(a)** Explanation tree and **(b)** causal explanation tree for the Academe network when the FinalMark is fail.

The ET method incorrectly selects Practice as the most important variable; Theory has a higher impact on the final mark according to this model. Again, it is because the ET method measures the importance of a target variable using its mutual information with other target variables, not with the evidence variable. Recall that it made bad choices for the Academe and Vacation networks as well.

The CET method made a more sensible choice by selecting Theory as the most important variable. There are three equally good explanations according to this method: (*bad theory*), (*average theory*, *bad practice*), and (*good theory*, *bad practice*). The reason that this method cannot distinguish these explanations is that it marks the branches using the log ratio between the posterior and prior probabilities of the evidence, which is proportional to the likelihood measure.

### 5.4 Asia

The Asia network is first introduced by Lauritzen and Spiegelhalter (1988) and was used by Nielsen et al. (2008) to discuss the CET method. The probabilities of this network are parameterized as follows.

$$P(VisitToAsia = yes) = 0.01;$$
$$P(Smoking = yes) = 0.5;$$
$$P(Tuberculosis = yes | VisitToAsia = yes) = 0.05;$$

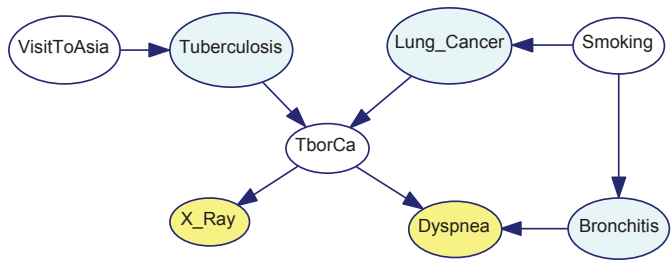

Figure 10: The Asia network.

$$P(Tuberculosis = yes | VisitToAsia = no) = 0.01;$$
$$P(LungCancer = yes | Smoking = yes) = 0.1;$$
$$P(LungCancer = yes | Smoking = no) = 0.01;$$
$$P(Bronchitis = yes | Smoking = yes) = 0.6;$$
$$P(Bronchitis = yes | Smoking = no) = 0.3;$$
$$P(TborCa = yes | Tuberculois = yes \lor LungCancer = yes) = 1.0;$$
$$P(TborCa = no | Tuberculois = no, LungCancer = no) = 0.0;$$
$$P(Dyspnea = yes | TborCa = yes, Bronchitis = yes) = 0.9;$$
$$P(Dyspnea = yes | TborCa = yes, Bronchitis = no) = 0.7;$$
$$P(Dyspnea = yes | TborCa = no, Bronchitis = yes) = 0.8;$$
$$P(Dyspnea = yes | TborCa = no, Bronchitis = no) = 0.1;$$
$$P(X\_ray = abnormal | TborCa = yes) = 0.98;$$
$$P(X\_ray = abnormal | TborCa = no) = 0.05;$$

We consider two different observations in the Asia network: Dyspnea is present, or X-ray is abnormal. Table 7 and Figure 11 show the explanations found by the various methods for these two observations using the target variable set $\{Bronchitis, LungCancer, Tuberculosis\}$. It is interesting that K-MRE obtained quite intuitive and concise top explanations for both observations: ($Bronchitis$), ($LungCancer$), or ($Tuberculosis$). Here we use a disease name such as $Bronchitis$ to denote the presence of the disease and a negation such as $\neg Bronchitis$ to denote the absence of the disease. The ranking of the explanations, however, is different. For the first observation, ($Bronchitis$) is a better explanation for Dyspnea than both ($Tuberculosis$) and ($LungCancer$) because the conditional probabilities show that the presence of Bronchitis has a larger effect on Dyspnea. For the second observation, Tuberculosis and LungCancer are the ancestors of X-ray, while Bronchitis has no direct effect on X-ray and only receives a small GBF. The reason that ($Bronchitis$)'s GBF is still greater than 1.0 is that Bronchitis increases the likelihood of Smoking, which in turn increases the likelihood of abnormal X-ray.

K-MAP identified Bronchitis as part of the best explanation for the first observation, but it included both Bronchitis and LungCancer as part of the best explanation for the second observation, even though Bronchitis is not a direct contributor to the abnormal

| Asia | Explanations | Scores |
|------|-------------|--------|
| | Dyspnea present | |
| K-MRE | $(Bronchitis)$ | 6.1391 |
| | $(LungCancer)$ | 1.9678 |
| | $(Tuberculosis)$ | 1.8276 |
| K-MAP | $(\neg LungCancer, \neg Tuberculosis, Bronchitis)$ | 0.3313 |
| | $(\neg LungCancer, \neg Tuberculosis, \neg Bronchitis)$ | 0.0521 |
| | $(LungCancer, \neg Tuberculosis, Bronchitis)$ | 0.0521 |
| K-SIMP | $(\neg LungCancer, \neg Bronchitis)$ | 0.9000 |
| | $(Bronchitis)$ | 0.8080 |
| | $(\neg Tuberculosis)$ | 0.4323 |
| | Abnormal X-ray | |
| K-MRE | $(LungCancer)$ | 16.4231 |
| | $(Tuberculosis)$ | 9.6886 |
| | $(Bronchitis)$ | 1.2535 |
| K-MAP | $(LungCancer, \neg Tuberculosis, Bronchitis)$ | 0.0305 |
| | $(\neg LungCancer, \neg Tuberculosis, Bronchitis)$ | 0.0261 |
| | $(LungCancer, \neg Tuberculosis, \neg Bronchitis)$ | 0.0228 |
| K-SIMP | $(\neg LungCancer)$ | 0.9800 |
| | $(\neg Tuberculosis)$ | 0.1012 |

Table 7: Top explanations found by K-MRE, K-MAP, and K-SIMP for the Asia network given two different observations.

X-ray. The second best explanation for the first observation claims none of the diseases is present, which is clearly not a good explanation.

The K-SIMP method found quite perplexing explanations for the observations. The best explanation is $(\neg LungCancer, \neg Bronchitis)$ for the first observation and $(\neg LungCancer)$ for the second observation. The reason that K-SIMP did not find good explanations for this network is that K-SIMP is restricted to the solutions found by K-MAP. The method was still able to find $(Bronchitis)$ as the second best explanation for the first observation, but it missed altogether for the second observation.

The ET method was able to find the most important variables in explaining both observations, but it fell short in recognizing the importance of Tuberculosis in explaining the second observation. It only expanded LungCancer in explaining the first observation.

The CET method was able to find intuitive explanations. The best explanations are $(Bronchitis)$ for Dyspnea and $(LungCancer)$ for abnormal X-ray. However, one drawback of its explanation tree is that the tree structure requires each explanation to start from the root. For example, one of the explanations for explaining Dyspnea is $(\neg Bronchitis, \neg Lung$-$Cancer, Tuberculosis)$. It is arguable that $(Tuberculosis)$ itself is a good explanation for explaining Dyspnea; it is not really necessary to include $\neg Bronchitis$ and $\neg LungCancer$ as part of the explanation. We believe that this is another common drawback of the two explanation tree methods caused because of the tree representation that they use.

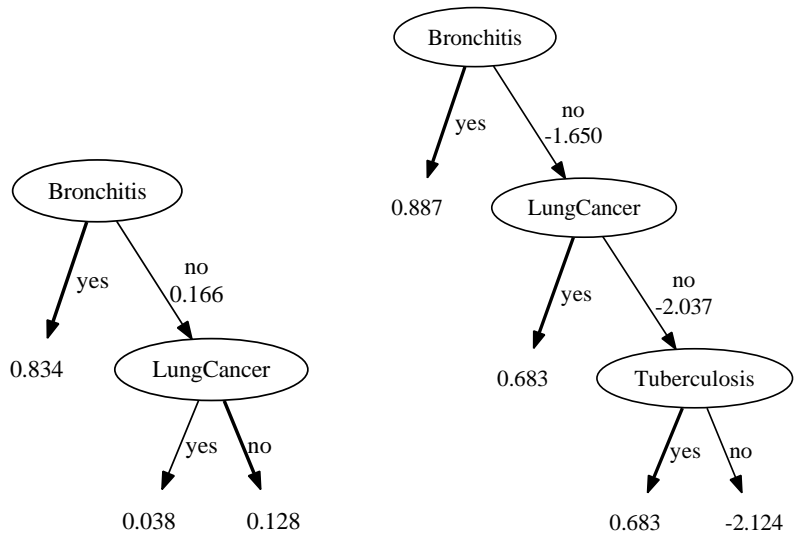

**(a)** ET for Dyspnea          **(b)** CET for Dyspnea

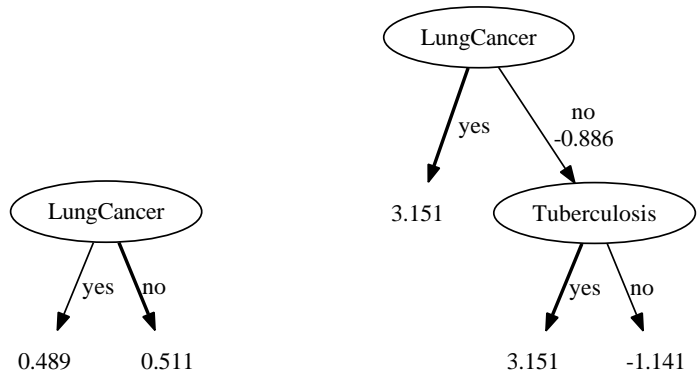

**(c)** ET for abnormal X-ray          **(d)** CET for abnormal X-ray

Figure 11: Explanation trees (ET) and causal explanation trees (CET) found for the Asia network given two different observations.

## 5.5 Circuit2

Model-based diagnosis is an application of abductive inference in Horn-clause logic theories (Peirce, 1948; de Kleer & Williams, 1987), which tries to find a minimal set of assumptions that, together with background knowledge, logically entail the observations that need explanation. However, methods for model-based diagnosis are developed based on logic theories. Entailment is either true or false for logic systems. These methods cannot be easily generalized to probabilistic expert systems such as Bayesian networks. In contrast, MRE can be easily applied to model-based diagnostic systems in which the faulty behaviors of the components are also specified.

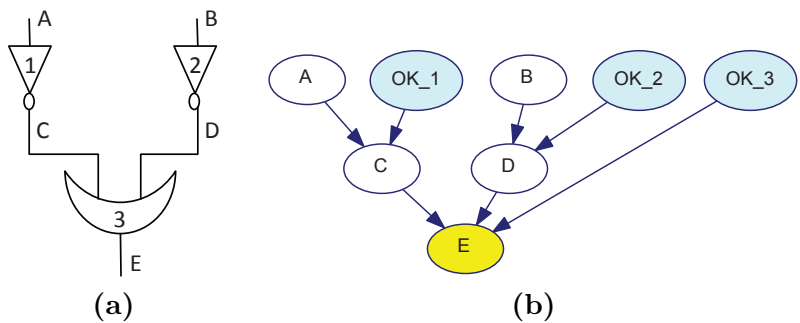

Figure 12: **(a)** A digital circuit from Darwiche (2009) and **(b)** its corresponding Bayesian network.

| Circuit in Figure 12 | Explanations | Scores |
| --- | --- | --- |
| K-MRE | $(\neg OK_3)$ | 4.0000 |
| | $(\neg OK_1, \neg OK_2)$ | 2.0000 |
| K-MAP | $(\neg OK_1, \neg OK_2, \neg OK_3)$ | 0.1250 |
| | $(\neg OK_1, OK_2, \neg OK_3)$ | 0.1250 |
| | $(OK_1, \neg OK_2, \neg OK_3)$ | 0.1250 |
| K-SIMP | $(\neg OK_3)$ | 1.0000 |
| | $(\neg OK_1, \neg OK_2)$ | 1.0000 |

Table 8: Top explanations found by K-MRE, K-MAP, and K-SIMP for the Circuit2 network given that the output is observed to be low.

We consider the digital circuit in Figure 12(a) used by Darwiche (2009) to discuss methods for model-based diagnosis. Gates 1 and 2 are inverters, and gate 3 is an OR gate. The prior probability that each of the gates is abnormal is 0.5. When an inverter is abnormal, it outputs low when the input is low, and outputs high with probability 0.5 when the input is high. When an OR gate is abnormal, it always outputs low. This digital circuit can be modeled as the Bayesian network in Figure 12(b). We consider the case when output E is observed to be low. The two kernel model-based diagnoses are $(\neg OK_1, \neg OK_2)$ and $(\neg OK_3)$. Table 8 and Figure 13 show the explanations found by the various methods for the observation using the target variable set $\{OK_1, OK_2, OK_3\}$.

K-MRE was able to find the two kernel diagnoses as its top two explanations, with $(\neg OK_3)$ receives a higher GBF than $(\neg OK_1, \neg OK_2)$. The higher GBF score is due to $(\neg OK_3)$'s higher prior and posterior probabilities than $(\neg OK_1, \neg OK_2)$. In comparison, methods for model-based diagnosis typically treat the two explanations as equally good.

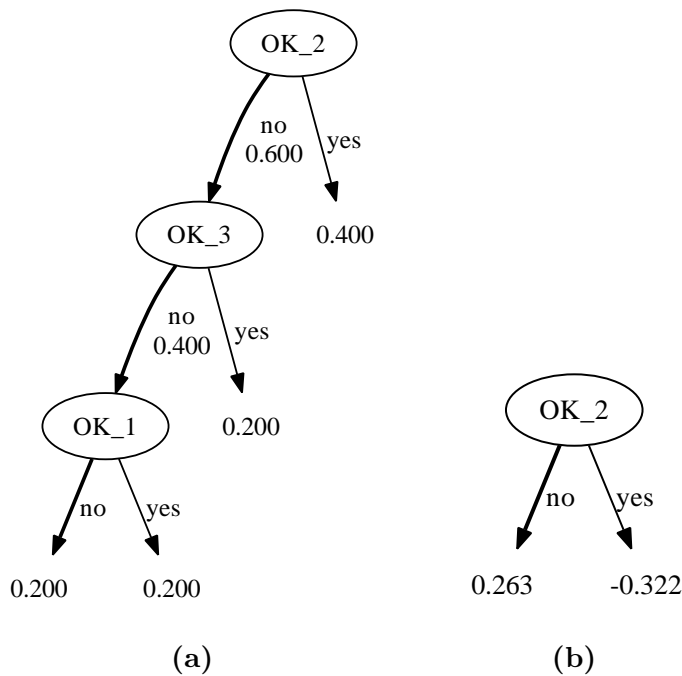

**(a)**            **(b)**

Figure 13: **(a)** Explanation tree and **(b)** causal explanation tree for the Circuit2 network given that the output is observed to be low.

K-MAP was not able to single out the two kernel diagnoses. In fact, many MAP solutions have the same posterior probability, including the explanation in which all the gates are defective.

The K-SIMP method found the same two explanations as K-MRE. Therefore, the simplification method helped in simplifying the explanations in this network. However, K-SIMP was not able to rank the two explanations either.

The two explanation tree methods completely misfired on this network. It is unclear how to make sense of the explanation tree by the ET method in Figure 13(a). The causal explanation tree in Figure 13(b) only expanded variable OK_2 and failed to find any of the kernel diagnoses. Again, it is because the explanation tree methods are greedy search methods and cannot recognize the compound effect of multiple variables.

## 5.6 Summary of the Case Studies

The case studies show that K-MRE was able to identify the most relevant target variables and find more concise and intuitive explanations than the other methods. GBF seems to be a plausible measure of explanatory power that can achieve both preciseness and conciseness in explanations at the same time. Another advantage of GBF is that we can use Table 1 as a general guidance for determining the significance of the explanations found by K-MRE. In contrast, probability seems not to be a good measure of explanatory power. Methods based on probability, such as K-MAP, are quite sensitive to modeling choices; they also lack the capability to indicate the most important parts of their explanations. The K-MAP simplification method was shown often to be able to simplify the solutions of K-MAP to get more concise explanations. But its fundamental drawback is that it is restricted to the solutions found by K-MAP and may not be able to find the best explanations. Also, it may reduce multiple top MAP solutions to the same explanation. The ET method uses the mutual information between the target variables as the criterion to select target variables to explain the evidence. The criterion was shown not to be very effective. Also, using probability to rank the explanations makes the ET method very sensitive to the user-specified threshold value for bounding the probabilities of the branches. The CET method is good at identifying individual target variables that are important, but it often fails to recognize the significant compound effect of multiple variables. It is because the CET method, as well as the ET method, is based on a greedy search that only considers one variable at a time. Another common drawback of the explanation tree methods is that the tree representation fundamentally limits the capability of these methods to find concise explanations.

## 6. Concluding Remarks

In this paper, we introduced the Most Relevant Explanation (MRE) method for finding explanations for given evidence in Bayesian networks. Our study shows that MRE has several desirable theoretical properties that enable MRE to automatically identify the most relevant target variables to find an explanation for the evidence. MRE is also able to capture the unique explaining-away phenomenon often represented in Bayesian networks. We defined two dominance relations among the MRE solutions and used them to develop a K-MRE method to find a set of top MRE solutions that is both diverse and representative. The results of the case studies on a set of benchmark Bayesian networks agree quite well with our theoretical understandings of MRE. Another contribution of this research is that it also made clear the properties and drawbacks of several existing relevance measures and explanation methods.

## Acknowledgments

This research was supported by the National Science Foundation grants IIS-0842480, IIS-0953723, and EPS-0903787. Part of this research has previously been presented in DX-07 (Yuan & Lu, 2007), AAAI-08 (Yuan & Lu, 2008), UAI-09 (Yuan, Liu, Lu, & Lim, 2009), and ExaCt-09 (Yuan, 2009). We thank the editors and anonymous reviewers for their constructive comments.

## Appendix A. Proofs

The following are the proofs of the theorems and corollaries.

## A.1 Proof of Theorem 1

**Proof:** The likelihood measure can be expressed as

$$P(\mathbf{e}|\mathbf{x}) = \frac{P(\mathbf{x}|\mathbf{e})P(\mathbf{e})}{P(\mathbf{x})} = r(\mathbf{x};\mathbf{e})P(\mathbf{e}).$$

Therefore, a fixed likelihood $P(\mathbf{e}|\mathbf{x})$ indicates that the belief update ratio $r(\mathbf{x};\mathbf{e})$ remains constant while the prior and posterior probabilities may vary. Furthermore, GBF can be expressed as follows.

$$GBF(\mathbf{x};\mathbf{e}) = 1 + \frac{r(\mathbf{x};\mathbf{e}) - 1}{1 - r(\mathbf{x};\mathbf{e})P(\mathbf{x})}.$$

Therefore, GBF is monotonically non-decreasing for a fixed belief update ratio $r(\mathbf{x};\mathbf{e})$ greater than or equal to 1.0 as the prior and posterior probabilities increase.

$\square$

## A.2 Proof of Theorem 2

**Proof:**

$$
\begin{aligned}
GBF(\mathbf{x},\mathbf{y};\mathbf{e}) &= \frac{P(\mathbf{x},\mathbf{y}|\mathbf{e})(1 - P(\mathbf{x},\mathbf{y}))}{P(\mathbf{x},\mathbf{y})(1 - P(\mathbf{x},\mathbf{y}|\mathbf{e}))} \\
&= \frac{P(\mathbf{x}|\mathbf{e})P(\mathbf{y}|\mathbf{x},\mathbf{e})(1 - P(\mathbf{y}|\mathbf{x})P(\mathbf{x}))}{P(\mathbf{x})P(\mathbf{y}|\mathbf{x})(1 - P(\mathbf{y}|\mathbf{x},\mathbf{e})P(\mathbf{x}|\mathbf{e}))} \\
&= \frac{P(\mathbf{x}|\mathbf{e})}{P(\mathbf{x})} \frac{1 - P(\mathbf{x}) + \frac{1}{P(\mathbf{y}|\mathbf{x})} - 1}{1 - P(\mathbf{x}|\mathbf{e}) + \frac{1}{P(\mathbf{y}|\mathbf{x},\mathbf{e})} - 1}
\end{aligned}
$$

The above equation is less than or equal to $GBF(\mathbf{x};\mathbf{e})$ when

$$
\begin{aligned}
\frac{\frac{1}{P(\mathbf{y}|\mathbf{x})} - 1}{\frac{1}{P(\mathbf{y}|\mathbf{x},\mathbf{e})} - 1} &\leq \frac{1 - P(\mathbf{x})}{1 - P(\mathbf{x}|\mathbf{e})} \\
\Leftrightarrow \frac{P(\mathbf{y}|\mathbf{x},\mathbf{e})(1 - P(\mathbf{y}|\mathbf{x}))}{P(\mathbf{y}|\mathbf{x})(1 - P(\mathbf{y}|\mathbf{x},\mathbf{e}))} &\leq \frac{P(\overline{\mathbf{x}})}{P(\overline{\mathbf{x}}|\mathbf{e})} \\
\Leftrightarrow CBF(\mathbf{y};\mathbf{e}|\mathbf{x}) &\leq \frac{1}{r(\overline{\mathbf{x}};\mathbf{e})}.
\end{aligned}
$$

$\square$

## A.3 Proof of Corollary 1

**Proof:** The corollary follows from Theorem 2. Here we present another way to prove it.

$$
\begin{aligned}
GBF(\mathbf{x},y;\mathbf{e}) &= \frac{P(\mathbf{x},y|\mathbf{e})(1 - P(\mathbf{x},y))}{P(\mathbf{x},y)(1 - P(\mathbf{x},y|\mathbf{e}))} \\
&= \frac{P(\mathbf{x}|\mathbf{e})P(y)(1 - P(y)P(\mathbf{x}))}{P(\mathbf{x})P(y)(1 - P(y)P(\mathbf{x}|\mathbf{e}))} \\
&= \frac{P(\mathbf{x}|\mathbf{e})(1 - P(y)P(\mathbf{x}))}{P(\mathbf{x})(1 - P(y)P(\mathbf{x}|\mathbf{e}))}.
\end{aligned}
$$

Because $P(\mathbf{x}|\mathbf{e}) > P(\mathbf{x})$, we have

$$
\begin{aligned}
GBF(\mathbf{x}, y; \mathbf{e}) &= \frac{P(\mathbf{x}|\mathbf{e})(1 - P(y)P(\mathbf{x}))}{P(\mathbf{x})(1 - P(y)P(\mathbf{x}|\mathbf{e}))} \\
&= \frac{P(\mathbf{x}|\mathbf{e})(1 - P(\mathbf{x}) + (1 - P(y))P(\mathbf{x}))}{P(\mathbf{x})(1 - P(\mathbf{x}|\mathbf{e}) + (1 - P(y))P(\mathbf{x}|\mathbf{e}))} \\
&< \frac{P(\mathbf{x}|\mathbf{e})(1 - P(\mathbf{x}) + (1 - P(y))P(\mathbf{x}))}{P(\mathbf{x})(1 - P(\mathbf{x}|\mathbf{e}) + (1 - P(y))P(\mathbf{x}))} \\
&\leq \frac{P(\mathbf{x}|\mathbf{e})(1 - P(\mathbf{x}))}{P(\mathbf{x})(1 - P(\mathbf{x}|\mathbf{e}))} \\
&= GBF(\mathbf{x}; \mathbf{e}) \ .
\end{aligned}
$$

$\square$

### A.4 Proof of Corollary 2

**Proof:** This corollary can be proved in a similar way as in Corollary 1.

$$
\begin{aligned}
GBF(\mathbf{x}, y; \mathbf{e}) &= \frac{P(\mathbf{x}, y|\mathbf{e})(1 - P(\mathbf{x}, y))}{P(\mathbf{x}, y)(1 - P(\mathbf{x}, y|\mathbf{e}))} \\
&= \frac{P(\mathbf{x}|\mathbf{e})P(y|\mathbf{x}, \mathbf{e})(1 - P(y|\mathbf{x})P(\mathbf{x}))}{P(\mathbf{x})P(y|\mathbf{x})(1 - P(y|\mathbf{x}, \mathbf{e})P(\mathbf{x}|\mathbf{e}))} \\
&= \frac{P(\mathbf{x}|\mathbf{e})P(y|\mathbf{x})(1 - P(y|\mathbf{x})P(\mathbf{x}))}{P(\mathbf{x})P(y|\mathbf{x})(1 - P(y|\mathbf{x})P(\mathbf{x}|\mathbf{e}))} \\
&= \frac{P(\mathbf{x}|\mathbf{e})(1 - P(y|\mathbf{x})P(\mathbf{x}))}{P(\mathbf{x})(1 - P(y|\mathbf{x})P(\mathbf{x}|\mathbf{e}))} .
\end{aligned}
$$

Because $P(\mathbf{x}|\mathbf{e}) > P(\mathbf{x})$, we have

$$
\begin{aligned}
GBF(\mathbf{x}, y; \mathbf{e}) &= \frac{P(\mathbf{x}|\mathbf{e})(1 - P(y|\mathbf{x})P(\mathbf{x}))}{P(\mathbf{x})(1 - P(y|\mathbf{x})P(\mathbf{x}|\mathbf{e}))} \\
&= \frac{P(\mathbf{x}|\mathbf{e})(1 - P(\mathbf{x}) + (1 - p(y|\mathbf{x}))P(\mathbf{x}))}{P(\mathbf{x})(1 - P(\mathbf{x}|\mathbf{e}) + (1 - p(y|\mathbf{x}))P(\mathbf{x}|\mathbf{e}))} \\
&< \frac{P(\mathbf{x}|\mathbf{e})(1 - P(\mathbf{x}) + (1 - p(y|\mathbf{x}))P(\mathbf{x}))}{P(\mathbf{x})(1 - P(\mathbf{x}|\mathbf{e}) + (1 - p(y|\mathbf{x}))P(\mathbf{x}))} \\
&\leq \frac{P(\mathbf{x}|\mathbf{e})(1 - P(\mathbf{x}))}{P(\mathbf{x})(1 - P(\mathbf{x}|\mathbf{e}))} \\
&= GBF(\mathbf{x}; \mathbf{e}) \ .
\end{aligned}
$$

$\square$

### A.5 Proof of Corollary 3

**Proof:**

$$
GBF(\mathbf{x}, y; \mathbf{e}) = \frac{P(\mathbf{x}, y|\mathbf{e})(1 - P(\mathbf{x}, y))}{P(\mathbf{x}, y)(1 - P(\mathbf{x}, y|\mathbf{e}))} = \frac{P(\mathbf{x}|\mathbf{e})P(y|\mathbf{x}, \mathbf{e})(1 - P(y|\mathbf{x})P(\mathbf{x}))}{P(\mathbf{x})P(y|\mathbf{x})(1 - P(y|\mathbf{x}, \mathbf{e})P(\mathbf{x}|\mathbf{e}))} .
$$

Because $P(y|\mathbf{x}, \mathbf{e}) \leq P(y|\mathbf{x})$,

$$GBF(\mathbf{x}, y; \mathbf{e}) \leq \frac{P(\mathbf{x}|\mathbf{e})P(y|\mathbf{x})(1 - P(y|\mathbf{x})P(\mathbf{x}))}{P(\mathbf{x})P(y|\mathbf{x})(1 - P(y|\mathbf{x})P(\mathbf{x}|\mathbf{e}))} = \frac{P(\mathbf{x}|\mathbf{e})(1 - P(y|\mathbf{x})P(\mathbf{x}))}{P(\mathbf{x})(1 - P(y|\mathbf{x})P(\mathbf{x}|\mathbf{e}))}.$$

Because $P(\mathbf{x}|\mathbf{e}) > P(\mathbf{x})$, we have

$$
\begin{aligned}
GBF(\mathbf{x}, y; \mathbf{e}) &= \frac{P(\mathbf{x}|\mathbf{e})(1 - P(y|\mathbf{x})P(\mathbf{x}))}{P(\mathbf{x})(1 - P(y|\mathbf{x})P(\mathbf{x}|\mathbf{e}))} \\
&= \frac{P(\mathbf{x}|\mathbf{e})(1 - P(\mathbf{x}) + (1 - p(y|\mathbf{x}))P(\mathbf{x}))}{P(\mathbf{x})(1 - P(\mathbf{x}|\mathbf{e}) + (1 - p(y|\mathbf{x}))P(\mathbf{x}|\mathbf{e}))} \\
&< \frac{P(\mathbf{x}|\mathbf{e})(1 - P(\mathbf{x}) + (1 - p(y|\mathbf{x}))P(\mathbf{x}))}{P(\mathbf{x})(1 - P(\mathbf{x}|\mathbf{e}) + (1 - p(y|\mathbf{x}))P(\mathbf{x}))} \\
&\leq \frac{P(\mathbf{x}|\mathbf{e})(1 - P(\mathbf{x}))}{P(\mathbf{x})(1 - P(\mathbf{x}|\mathbf{e}))} \\
&= GBF(\mathbf{x}; \mathbf{e}) \ .
\end{aligned}
$$

$\square$

## A.6 Proof of Theorem 3

**Proof:** When $C$ is not observed, we have the following equality.

$$P(B|A, \mathbf{x}, \mathbf{y}) = P(B|\mathbf{x}, \mathbf{y}) = P(B|\neg A, \mathbf{x}, \mathbf{y}) = P(B|\mathbf{y}). \tag{19}$$

Therefore, we have

$$P(B|A, c, \mathbf{x}, \mathbf{y}) \leq P(B|c, \mathbf{x}, \mathbf{y}) \leq P(B|\neg A, c, \mathbf{x}, \mathbf{y})$$

$$\Leftrightarrow \quad 1 - P(B|\neg A, c, \mathbf{x}, \mathbf{y}) \leq 1 - P(B|c, \mathbf{x}, \mathbf{y}) \leq 1 - P(B|A, c, \mathbf{x}, \mathbf{y})$$

$$\Leftrightarrow \quad \frac{P(B|A,c,\mathbf{x},\mathbf{y})}{1 - P(B|A,c,\mathbf{x},\mathbf{y})} \leq \frac{P(B|c,\mathbf{x},\mathbf{y})}{1 - P(B|c,\mathbf{x},\mathbf{y})} \leq \frac{P(B|\neg A,c,\mathbf{x},\mathbf{y})}{1 - P(B|\neg A,c,\mathbf{x},\mathbf{y})}$$

$$\Leftrightarrow \quad \frac{P(B|A,c,\mathbf{x},\mathbf{y})}{1 - P(B|A,c,\mathbf{x},\mathbf{y})} \frac{1 - P(B|A,\mathbf{x},\mathbf{y})}{P(B|A,\mathbf{x},\mathbf{y})} \leq \frac{P(B|c,\mathbf{x},\mathbf{y})}{1 - P(B|c,\mathbf{x},\mathbf{y})} \frac{1 - P(B|\mathbf{x},\mathbf{y})}{P(B|\mathbf{x},\mathbf{y})} \leq \frac{P(B|\neg A,c,\mathbf{x},\mathbf{y})}{1 - P(B|\neg A,c,\mathbf{x},\mathbf{y})} \frac{1 - P(B|\neg A,\mathbf{x},\mathbf{y})}{P(B|\neg A,\mathbf{x},\mathbf{y})}$$

$$\Leftrightarrow \quad GBF(B; c|A, \mathbf{x}, \mathbf{y}) \leq GBF(B; c|\mathbf{x}, \mathbf{y}) \leq GBF(B; c|\neg A, \mathbf{x}, \mathbf{y}) \qquad .$$

$\square$